\pdfoutput=1

\documentclass[11pt]{article}

\usepackage[table]{xcolor}
\usepackage{enumitem}
\usepackage{threeparttable}

\usepackage{acl} %

\usepackage{times}
\usepackage{latexsym}

\usepackage[T1]{fontenc}

\usepackage[utf8]{inputenc}
\usepackage{tikz}

\usepackage{microtype}

\usepackage{inconsolata}

\usepackage{dirtree}
\usepackage{amsmath}
\usepackage{amsthm}
\usepackage{booktabs}
\usepackage{tabularx}

\usepackage{comment}
\usepackage{mdframed}
\usepackage{amssymb}
\usepackage{subcaption}
\usepackage{float} 
\usepackage[normalem]{ulem}
\usepackage{tikz}
\usetikzlibrary{positioning}

\usetikzlibrary{arrows.meta}
\usepackage[edges]{forest}
\usepackage{fontawesome}
\usepackage[most]{tcolorbox}
\usepackage{multirow}
\usepackage{etoolbox} %
\definecolor{fallacy}{HTML}{EA9999}
\definecolor{credibility}{HTML}{6AA84F}
\definecolor{logic}{HTML}{E69138}
\definecolor{emotion}{HTML}{3D85C6}
\definecolor{mycustomcolor}{HTML}{1F77B4}
\definecolor{defcolor}{HTML}{6AA84F}

\definecolor{example1}{HTML}{63BCFF}
\definecolor{example2}{HTML}{FF6376}
\definecolor{example3}{HTML}{63FF83}
\definecolor{example4}{HTML}{F5AD64}

\tikzset{
  level 1/.style={sibling distance=35mm, level distance=15mm},
  level 2/.style={sibling distance=30mm},
  level 3/.style={sibling distance=13mm}
}

\newtcolorbox[auto counter,number within=section]{mydef}[2][]{
    colback=white,
    colframe=defcolor,
    colbacktitle=defcolor,
    coltitle=white,
    title=Definition~\thetcbcounter: #2,
    enhanced,
    attach boxed title to top left={yshift=-2.3mm, xshift=2mm},
    #1
}
\newtcolorbox[auto counter,number within=section]{myexamplebox}[1][]{
    colback=white,
    colframe=mycustomcolor,
    colbacktitle=mycustomcolor,
    title=Example~\thetcbcounter,
    enhanced,
    attach boxed title to top left={yshift=-2.3mm, xshift=2mm},
    #1
}

\definecolor{softblue}{rgb}{0.88, 0.95, 0.97} %
\newcommand{\fallacy}[5]{%
    \paragraph{#1}
    \begin{itemize}
        \item[] \textbf{Informal:} #2
        \item[] \textbf{Formal:} #3
        \item[] \textbf{Example:} #4
        \item[] \textbf{Annotation with Variables:} #5
    \end{itemize}
}
\newmdenv[
    topline=false,
    bottomline=false,
    rightline=false,
    linecolor=lightgray,
    linewidth=4pt,
    innertopmargin=1pt,
    innerbottommargin=1pt,
    innerleftmargin=5pt,
    innerrightmargin=0pt,
    skipabove=5pt,
    skipbelow=5pt
  ]{siderules}
  
\newtheorem{proposition}{Proposition}[section]

\newcommand{\adhominem}{\emph{abusive ad hominem}}

\newcommand{\tuquoque}{\emph{tu quoque}}
\newcommand{\guiltbyassociation}{\emph{guilt by association}}
\newcommand{\appealtopity}{\emph{appeal to pity}}
\newcommand{\strawman}{\emph{strawman}}
\newcommand{\appealtoworseproblem}{\emph{appeal to worse problem}}
\newcommand{\appealtoridicule}{\emph{appeal to ridicule}}

\usepackage{amssymb}

\usepackage{mathtools}

\usepackage{subcaption}

\usepackage{rotating}
\usepackage{nicematrix}

\title{MAFALDA\@: A Benchmark and Comprehensive Study\\ of Fallacy Detection and Classification}
\author{
    Chadi Helwe\textsuperscript{\rm 1},
    Tom Calamai\textsuperscript{\rm 1, 2},
    Pierre-Henri Paris\textsuperscript{\rm 1},
    Chlo\'e Clavel\textsuperscript{\rm 3},
    and Fabian M. Suchanek\textsuperscript{\rm 1}\\
    \textsuperscript{\rm 1} Télécom Paris, Institut Polytechnique de Paris, France\\
    \textsuperscript{\rm 2} INRIA Saclay, Amundi, France\\
    \textsuperscript{\rm 3} INRIA Paris, France\\
    \texttt{firstName.lastName@telecom-paris.fr,firstName.lastName@inria.fr}\\
}

\begin{document}

\maketitle

\begin{abstract}
    We introduce MAFALDA, a benchmark for fallacy classification that merges and unites previous fallacy datasets. It comes with a taxonomy that aligns, refines, and unifies existing classifications of fallacies. We further provide a manual annotation of a part of the dataset together with manual explanations for each annotation. We propose a new annotation scheme tailored for subjective NLP tasks, and a new evaluation method designed to handle subjectivity.
We then evaluate several language models under a zero-shot learning setting and human performances on MAFALDA to assess their capability to detect and classify fallacies.

\end{abstract}

\section{Introduction}\label{sec:intro}

A fallacy is an erroneous or invalid way of reasoning. 
Consider, e.g.,~the argument ``\textit{You must either support my presidential candidacy or be against America!}''. 
This argument is a \emph{false dilemma} fallacy: it wrongly assumes no other alternatives.
\textbf{Fallacies can be found in various forms of communication}, including speeches, advertisements~\cite{danciu2014manipulative}, Twitter/X posts~\cite{macagno2022argumentation}, and political debates~\cite{balalauStageAudiencePropaganda2021,goffredoFallaciousArgumentClassification2022}. 
They are also part of propaganda techniques employed to shape public opinion and promote specific agendas.
Most notably, fallacies 
played a role in the 2016 Brexit referendum~\cite{zappettini2019brexit}, 
and the debate about COVID-19 vaccinations~\cite{elsayed2020fallacies}, where fake news spread on news outlets and in social networks~\cite{martinoFineGrainedAnalysisPropaganda2019, sahaiBreakingInvisibleWall2021,balalauStageAudiencePropaganda2021}.
Detecting and identifying these fallacies is thus a task of broad importance. 

The recent advances in deep learning and the availability of more data have given rise to approaches for detecting and classifying fallacies in text automatically~\cite{martinoFineGrainedAnalysisPropaganda2019,al-omari-etal-2019-justdeep,balalauStageAudiencePropaganda2021,sahaiBreakingInvisibleWall2021,abdullahDetectingPropagandaTechniques2022}. 
And yet, this work is fragmented: \textbf{most approaches focus on specific types of corpora} (e.g.,~only speeches) or \textbf{specific types of fallacies} (e.g.,~only \emph{ad hominem} fallacies).
Furthermore, \textbf{not all works use the same types of fallacies}, 
there is no consensus on a common terminology~\cite{sep-fallacies}, and fallacies come at different levels of granularity: an \emph{appeal to emotion} can be, for instance, an \emph{appeal to anger}, \emph{fear}, \emph{pride}, or \emph{pity}. 
Most importantly, \textbf{annotating fallacies is an inherently subjective task}. While previous works acknowledge the subjectivity, none explicitly embraces it. %
On the contrary, the annotators typically aim for a unique annotation -- by discussion or vote. 
Additionally, \textbf{existing works do not give human performances} on the benchmarks and evaluate only models.

This paper addresses these drawbacks by introducing the Multi-level Annotated Fallacy Dataset MAFALDA -- a manually created fallacy classification benchmark. %
Our contributions are: 
\begin{enumerate}[noitemsep,nolistsep,leftmargin=4mm]
    \item A taxonomy of fallacies that aligns, consolidates, and unifies %
    existing public fallacy collections (Section~\ref{sec:def}).
    \item A new annotation scheme -- coined disjunctive annotation scheme -- that accounts for the inherent subjectivity of fallacy annotation by permitting several correct annotations (Section~\ref{sec:annot_dataset}).
    \item A corpus that merges existing corpora, with 9,545 non-annotated texts and 200 manually annotated texts with 260 instances of fallacies, each with a manual justification (Section~\ref{sec:dataset}).
    \item A study of the performance of state-of-the-art language models and humans on our benchmark (Section~\ref{sec:xp}). 
\end{enumerate}

\noindent %
All our code and data are publicly available under a CC-BY-SA license\footnote{as imposed by the dataset from
\citet{goffredoFallaciousArgumentClassification2022}} at %
\url{https://github.com/ChadiHelwe/MAFALDA}, allowing our study to be reproduced and built upon.
We start our paper by discussing related work in Section~\ref{sec:related_work}.

\begin{figure*}
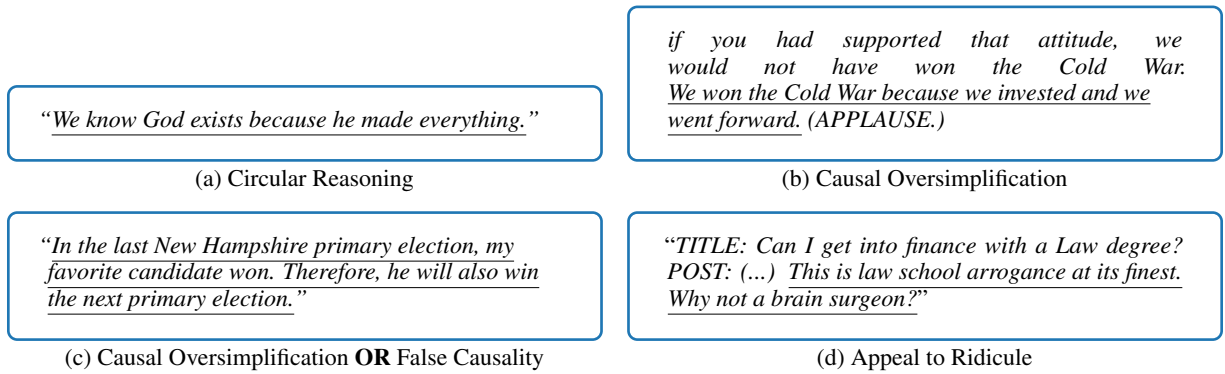

\centering
\footnotesize
\begin{minipage}[b]{.49\textwidth}
\centering
\begin{subfigure}[b]{\linewidth}
  \centering
  \begin{tcolorbox}[boxrule=1pt,colback=white,colframe=mycustomcolor] %
  \textit{``\underline{We know God exists because he made everything.}''}
  \end{tcolorbox}
  \vspace{-1em}
  \caption{Circular Reasoning}
  \label{ex:implicit}
\end{subfigure}
\end{minipage}%
\hfill %
\begin{minipage}[b]{.49\textwidth}
\centering
\begin{subfigure}[b]{\linewidth}
  \centering
  \begin{tcolorbox}[boxrule=1pt,colback=white,colframe=mycustomcolor]
  \textit{if you had supported that attitude, we would not have won the Cold War. \underline{We won the Cold War because we invested and we}\\\underline{went forward.} (APPLAUSE.)}
  \end{tcolorbox}
  \vspace{-1em}
  \caption{Causal Oversimplification}
  \label{ex:political}
\end{subfigure}
\end{minipage}

\vspace{0.2cm} %

\begin{minipage}[b]{.49\textwidth}
\begin{subfigure}[b]{\textwidth}
  \centering
  \begin{tcolorbox}[boxrule=1pt,colback=white,colframe=mycustomcolor]
  \textit{``\underline{In the last New Hampshire primary election, my} \underline{favorite candidate won. Therefore, he will also win} \underline{the next primary election.}''}
  \end{tcolorbox}
  \vspace{-1em}
  \caption{Causal Oversimplification \textbf{OR} False Causality}
  \label{ex:main}
\end{subfigure}
\end{minipage}
\hfill
\begin{minipage}[b]{.49\textwidth}
\begin{subfigure}[b]{\textwidth}
  \centering
  \begin{tcolorbox}[boxrule=1pt,colback=white,colframe=mycustomcolor]
``\textit{TITLE: Can I get into finance with a Law degree? POST: (...) \underline{This is law school arrogance at its finest.} \underline{Why not a brain surgeon?}}''
\end{tcolorbox}
  \vspace{-1em}
\caption{Appeal to Ridicule}
  \label{ex:finance_law_degree}
\end{subfigure}
\end{minipage}
\caption{Examples of Fallacies. The spans of the fallacies are \underline{underlined}. Example~\ref{ex:implicit} is from \citet{jinLogicalFallacyDetection2022}, \ref{ex:political}  from \citet{goffredoFallaciousArgumentClassification2022}, and \ref{ex:finance_law_degree} from \citet{sahaiBreakingInvisibleWall2021}. Detailed annotations are in Appendix~\ref{app:additional_examples}.
}
\label{fig:main}
\end{figure*}
\section{Related Work}\label{sec:related_work}

\subsection{Datasets}
Numerous works have created datasets of fallacies. 
\citet{habernalNameCallingDynamicsTriggers2018} created a dataset for \emph{ad hominem} fallacies from the ``Change My View'' subreddit. 
\citet{martinoFineGrainedAnalysisPropaganda2019} created a news article dataset featuring 18 fallacies such as \emph{red herring, appeal to authority, bandwagon, etc.}.
\citet{balalauStageAudiencePropaganda2021} trained models for propaganda technique identification using online forums.
\citet{sahaiBreakingInvisibleWall2021} compiled a Reddit-based corpus for fallacy detection with eight types of fallacies.
\citet{goffredoFallaciousArgumentClassification2022} introduced a dataset from American political debates with six different fallacy types. 
Along the same line, \citet{jinLogicalFallacyDetection2022} curated a claim dataset, containing 13 types of fallacies, based on online quizzes and the Climate Feedback website, employing a novel approach that mimics first-order logic. %
To address data annotation challenges, \citet{habernalArgotario2017} created the Argotario game for fallacy detection in QA pairs. It created a corpus of 5 fallacy types. 
In the domain of (dis/mis)information, \citet{musiCOVID2022} and \citet{alhindiMultitask2022} annotated fallacies in COVID-19 and climate change articles with ten types of fallacies.
Lastly, \citet{payandeh2023susceptible} developed LOGICOM to evaluate large language models' (LLMs) robustness against logical fallacies in debate scenarios.

While all of these works advanced the understanding of fallacy detection, 
the studied fallacies are not the same across different works and are sometimes outright disjoint.
The only work that creates a comprehensive taxonomy of fallacies is the (not yet peer-reviewed) work of \citet{hong2023closer}. However, this work enumerates 232 fallacies, which is clearly too many to be handled by a human. And indeed, their dataset is composed only of toy examples generated by GPT-4.

In this paper, we propose a benchmark that not only unifies public datasets on fallacy detection 
in a handy yet all-embracing taxonomy, but also comes with human annotations, human explanations, and evaluations for both language models and humans.

\subsection{Subjectivity and Annotation Challenges} \label{subsec:subjectivity}

Human label variation is inherently part of annotating complex and subjective tasks~\cite{plank2022problem}. It is usually addressed with strategies such as simplifying the task, majority votes, or reconciliation of discrepancies. \citet{goffredoFallaciousArgumentClassification2022} computed the Krippendorff's $\alpha$ on a subset of fallacies and reached inter-annotator agreements (IAAs) ranging from 0.46 to 0.60, which is a moderate agreement. 
On simpler tasks such as identifying only \emph{ad hominem} using two groups of 6 workers, \citet{habernalNameCallingDynamicsTriggers2018} reported a Cohen's $\kappa$ of 0.79, which is a good agreement. However, they acknowledge the difficulty of annotated sub-categories such as \tuquoque~and \guiltbyassociation~(they found a low IAA, but the value is not provided). 
When annotating spans of propaganda techniques, a complex task, \citet{martinoFineGrainedAnalysisPropaganda2019} found a $\gamma$ IAA of 0.26, which is low. However, they could increase the IAA up to 0.60 when adding a reconciliation step. 
In \citet{sahaiBreakingInvisibleWall2021}, the annotator had to identify one fallacy at a time, and they reached a Cohen's $\kappa$ of 0.515 (ranging from 0.38 to 0.64 based on the fallacy), which is a moderate agreement. They also computed the $\gamma$ for the span selection per fallacy type and found values between 0.60 to 0.80, which is a good agreement. This was expected since it is a binary classification task.
\citet{sahaiBreakingInvisibleWall2021,jinLogicalFallacyDetection2022,musiCOVID2022,alhindiMultitask2022} used a reconciliation step too to tackle discrepancies in the annotations.

In summary, IAA in related work is usually only moderate. Disagreements are interpreted as noise, and are removed with various strategies. 
In this paper, we propose not just to acknowledge the subjectivity of fallacy annotation but actually to follow through with it.
We contend that there are cases 
where \textbf{multiple, equally valid annotations can coexist for the same textual span}. 
Therefore, we propose a new subjective annotation scheme that allows for several alternative labels for the same span. %

\subsection{Taxonomies of Fallacies}

Logical fallacies have been studied and classified since the time of Aristotle.
There is a notable diversity in approaches and contents across various sources. 
The works of Aristotle (see \citet{SophisticalRefutations2023}) and \citet{whatelyElementsLogic1897}, despite their historical significance, present limitations in terms of the breadth of fallacies covered, listing only 13 fallacies each (our work, in contrast, finds more than 20).
\citet{downesStephenGuideLogical} offers a more extensive list with 36 fallacies. However, it still fails to mention common fallacies such as \emph{appeal to nature}, \emph{appeal to tradition}, and \emph{guilt by association}. 
\citet{fallacyfiles} provides an exhaustive list of 87 fallacies. Yet, it provides only a rudimentary hierarchy (classifying, e.g., \emph{no true Scotsman} as a sub-category of \emph{equivocation}). 
 \citet{logicalfallacies} lists 48 fallacies -- but lacks a hierarchical framework altogether. 
At the other end of the spectrum, \citet{encycl_philo}, \citet{bennettLogicallyFallaciousUltimate2012}, \citet{hong2023closer}, and \citet{wiki_fallacies} offer extensive compilations of 231, 300+, 232, and 149 fallacies respectively. Yet, such a sheer volume of fallacies would be challenging in practical annotation tasks, as the annotator would have to scan (or memorize) hundreds of different fallacies.

Our work, in contrast, is driven by today's practical application scenarios. It aims to systematize and classify the fallacies used in current works on fallacy annotation, detection, and classification.

\section{A Unified Taxonomy of Fallacies}\label{sec:def}

\subsection{Definitions}

We start with the definition of an argument, following \citet{copi2018introduction,britannica}:

\begin{mydef}[label={def:arg}]{Argument}
An argument consists of an assertion called the \emph{conclusion} and one or more assertions called \emph{premises}, where the premises are intended to establish the truth of the conclusion.
Premises or conclusions can be implicit in an argument.
\end{mydef}

\noindent Thus, an argument is typically of the form ``\textit{\textbf{Premise$_1$:} All humans are mortal. \textbf{Premise$_2$:} Socrates is human. \textbf{Conclusion:} Therefore, Socrates is mortal.}''. 
However, premises and conclusion can also appear in the opposite order and/or in the same sentence, as in ``\textit{Socrates is mortal because he is a human and all humans are mortal}''. 
In many real-world arguments, the premise and the conclusion are spread apart (as in ``\textit{Of course, Socrates is mortal! How can you doubt this? After all, he's human, and all humans are mortal!}''). Sometimes, premises are left implicit (as in ``\textit{Socrates is mortal because he is human}''). Even the conclusion can be implicit (as in ``\textit{Socrates is human and all humans are mortal}''). 
In the context of a discussion, an argument can attack another argument~\cite{dungAcceptabilityArgumentsIts1995}, in which case the conclusion is implicitly negated (``\textit{Socrates is immortal! -- But he is human!}'').

A valid argument is one where the truth of the premises guarantees the truth of the conclusion; otherwise, following \citet{copi2018introduction, britannica}, the argument is a fallacy:
\begin{mydef}[label={def:fallacy}]{Fallacy}
A fallacy is an argument where the premises do not entail the conclusion. %
\end{mydef}
\noindent We refer the reader to \citet{tina} for a discussion of a formal definition of textual entailment. Appendix~\ref{sec:edge} further distinguishes fallacies from other types of erroneous statements. %

\subsection{Taxonomy of Fallacies}\label{sec:tax}

\begin{figure*}
    \centering
    \includegraphics[width=1.0\textwidth]{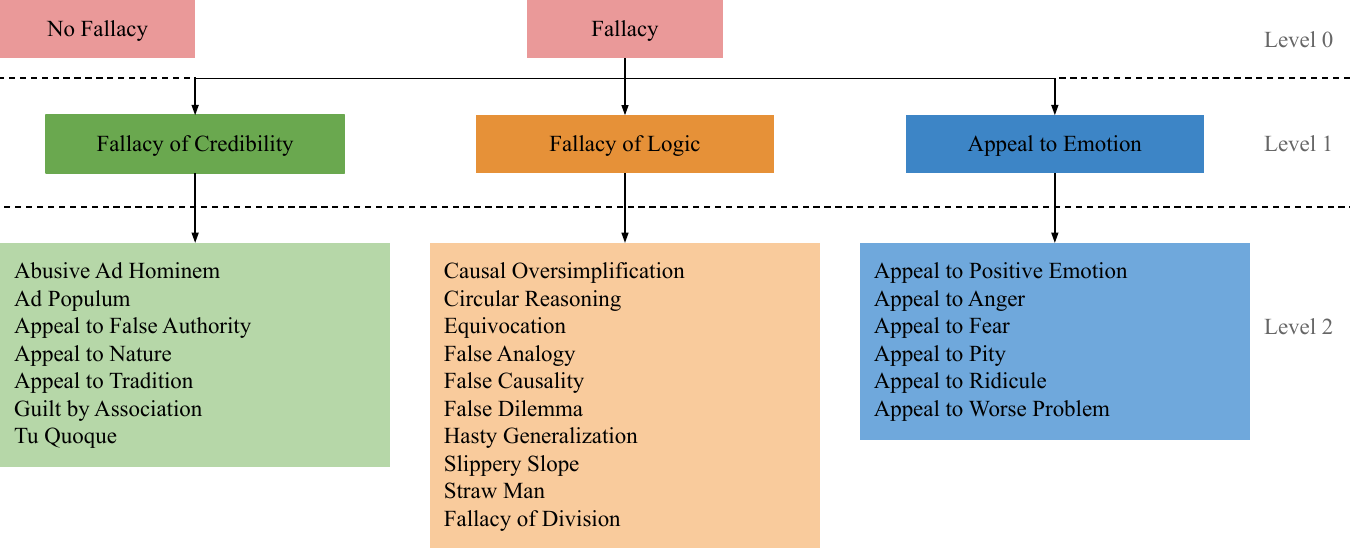}
    \caption{Tree structure of our taxonomy. Detailed definitions of the fallacies are in Appendix~\ref{app:informal_def}.
    }
    \label{fig:taxonomy}
\end{figure*}

In this paper, we propose a taxonomy that unifies and consolidates all types of fallacies used in current work on fallacy detection. 
We built our taxonomy manually, starting with a collection of fallacy types that are used in related work. 
Since the same fallacy can appear in different datasets under different names, we aligned equivalent fallacies manually. %
We used the definitions and guidelines in the source datasets to determine whether two fallacies are equivalent.
We removed fallacies that were too broad (e.g.,~\emph{appeal to emotion} could cover many emotions), fallacies that appeared in only a single work (e.g.,~\emph{confusion fallacy} appears only in~\citet{martinoFineGrainedAnalysisPropaganda2019}), and we merged fallacies that were too similar in their definitions (like \emph{begging the question} and \emph{circular reasoning}).
Some fallacies were not taken into account because they were not actually fallacies in our definition. These are, e.g., rhetorical techniques such as
\emph{flag waving} or \emph{repetition}.
Our list can obviously be extended in the future with new fallacies. Details on how our collection unifies existing works are in Appendix~\ref{app:comparison_of_types}.

We grouped our fallacies into broader categories to create a taxonomy on top of our collection. 
We chose the categories that Aristotle originally proposed~\cite{wisse1989ethos}, %
because %
it has been shown to be applicable across various forms of communication -- from political speeches to advertisements~\cite{wisse1989ethos}. This yields the following taxonomy:
\begin{enumerate}[noitemsep,nolistsep,leftmargin=4mm]
    \item \textbf{Level 0} is a binary classification, categorizing text as either fallacious or non-fallacious.
    \item \textbf{Level 1} groups fallacies into Aristotle's categories: `Pathos' (appeals to emotion), `Ethos' (fallacies of credibility), and `Logos' (fallacies of logic, relevance, or evidence).
    \item \textbf{Level 2} contains fine-grained fallacies within the broad categories of Level 1.
          For instance, under \emph{fallacy of credibility}, we have specific fallacies such as \emph{appeal to tradition}, \emph{ad populum}, and \emph{guilt by association}.
\end{enumerate}
\noindent Our taxonomy is shown in Figure~\ref{fig:taxonomy}.
For each fallacy, we provide \textbf{both a formal and an informal definition} in Appendix~\ref{app:informal_def} (inspired by \citet{bennettLogicallyFallaciousUltimate2012}).
For instance, the \appealtoridicule{} is informally defined as ``an argument that portrays the opponent's position as absurd or ridiculous with the intention of discrediting it.''. 
Formally, it is defined as ``$E_1$ claims $P$. $E_2$ makes $P$ look ridiculous, by misrepresenting $P$ ($P$'). Therefore, $\neg P$.'', 
where $E_i$ are entities (e.g., people, organizations, etc.), and $P$ is a proposition.

\section{Tackling Subjectivity in Annotations}\label{sec:annot_dataset}

\subsection{Subjectivity in Fallacy Annotation}\label{sec:subjectivity}

Annotating fallacies is an inherently subjective endeavor. 
To see this, consider Example (c) in Figure~\ref{fig:main}.
The argument goes that the candidate has to win again because he won last time. This can be seen as a \emph{false causality} fallacy: a cause-effect relationship is incorrectly inferred between two events that have nothing to do with each other.
However, it can also be seen as a \emph{causal oversimplification} fallacy. This is because
we can contend that having won the last election gives the candidate an edge over other candidates in terms of visibility, and thus makes it more likely that he wins this year's election as well. The argument is thus fallacious mainly because it fails to acknowledge other factors that play a role in re-election.

This simple example already shows subjectivity in fallacy annotations, where several annotations can be defended. It would be counter-productive if the annotators converged on, say, \emph{causal oversimplification}, so that every approach of fallacy annotation is penalized for proposing an (equally plausible) \emph{false causality}. There are other cases of legitimately differing opinions: One annotator may see implicit assertions that another annotator does not see. In ``\textit{Are you for America? Vote for me!}'' one reader may see the implicit ``\textit{or you must be against America}'' (which makes this a \emph{false dilemma}), while another annotator may see no such implicature. Annotators may also have different thresholds for fear (\emph{appeal to fear}) or insults (\emph{ad hominem}). 
Finally, different annotators have different background knowledge: A sentence such as ``\textit{Use disinfectants or you will get Covid-19!}'' 
may be read as a plausible warning by one annotator but as an \emph{appeal to fear} fallacy by an annotator who knows that Covid-19 does not spread via contaminated surfaces. 
We will now present a disjunctive annotation scheme that accounts for this inherent subjectivity.

\subsection{Disjunctive Annotation Scheme}\label{sec:annotation}
\label{sec:metrics}

Before presenting our annotation scheme, we need to establish some common ground:

\begin{mydef}[label={def:text}]{Text}
    A text is a sequence of sentences $st_1, \ldots, st_n$.
\end{mydef}

\noindent A \emph{span} on a text is a contiguous sequence of sentences. The set $S$ of all spans of a text $st_1, \ldots, st_n$ is thus $S=\{st_i \ldots st_j \mid 0<i\leq j\leq n\}$. 

\begin{mydef}[label={def:span}]{Span}
The span of a fallacy in a text is the smallest contiguous sequence of sentences that comprises the conclusion and the premises of the fallacy.
If the span comprises a pronoun that refers to a premise or to the conclusion, that premise or conclusion is not included in the span.
\end{mydef}

\noindent We work on the level of sentences, 
because previous work has shown that agreement on the token level is even harder to achieve~\cite{jinLogicalFallacyDetection2022}. 
We allow the use of pronouns to decrease the size of the spans: When a sentence refers to another sentence by a pronoun, that other sentence does not have to be part of the span.
For instance, in Example (d) of Figure~\ref{fig:main}, the premise of the fallacy is in the title of the post, and the conclusion is at the end of the text. Thus, a span that covers the entire fallacy would have to cover the entire post from title to end. %
However, the pronoun ``This'' refers to the title, and thus we can omit the title from the span.
Nevertheless, %
a span can comprise several sentences. 

A span can be annotated with a label (such as a fallacy type) by an annotator (or a group of them) or by a system.
We now propose the key element of our disjunctive annotation scheme, in which subjectivity is not projected away, but explicitly embraced by allowing for several equally valid labels for the same span. 

\begin{mydef}[label={def:gold_standard}]{Gold Standard}
Let $\mathcal{F}$ be the set of fallacy types and $\bot$ be a special label that means ``no fallacy''.

Given a text and its set of spans $S$, %
a gold standard $G$ is a set of pairs of a span $s \in S$ and a set of labels from $\mathcal{F}\cup\{\bot\}$: %
\[G \subseteq S \times (\mathcal{P}(\mathcal{F}\cup\{\bot\})\setminus\{\emptyset, \{\bot\}\})\] %
Here, $\mathcal{P}(\cdot)$ denotes the powerset. 

\end{mydef}

\noindent 
The gold standard associates a given span with one or more fallacy labels. If more than one label is present, this means that any label is acceptable as an annotation. The gold standard can also associate a span with $\bot$, which means that the annotation of this span is optional. However, in this case, the gold standard has to associate the span also with at least one other label,
as we are not interested in annotating non-fallacious sentences. The gold standard can also contain the same span twice, which means that the span has to be annotated with two labels. 
The alternative labels for a span
can be generated through various methods during the annotation process, e.g.,~one annotator giving alternatives, a group of annotators proposing different labels due to lack of consensus, or multiple independent annotators combining their labels (see Example~\ref{ex:anno}).

\noindent We define a prediction as the annotation of a text by a system or a user:

\begin{mydef}[label={def:prediction}]{Prediction}
Given a set of fallacy types $\mathcal{F}$, a text, and its set of spans $S$, a prediction $P$ is a set of pairs of a span $s\in S$ and a label $l\in\mathcal{F}$:
\[ P\subseteq S \times \mathcal{F}\] 
\end{mydef}
\noindent The following example gives substance to these definitions:
\begin{myexamplebox}[label={ex:anno}, breakable] %
Let ``$a\ b\ c\ d$'' be a text where $a$, $b$, $c$, and $d$ are sentences. 

Suppose \(S= \{a\ b, d\}\) (i.e.,~the sentences $a$ and $b$ are one fallacious span, and $d$ is a span of one fallacious sentence), $a\ b$ has labels $\{l_1, l_2\}$, and \(d\) has label \(\{l_3\}\).
In that case, \(G=\bigl\{ (a\ b, \{l_1, l_2\}) , (d, \{l_3\}) \bigl\}\)\\

An example of prediction $P$ could be $P=\{(a, l_1), (a, l_2), (b, l_3), (c, l_4),$ $(d, l_1)\}$
\end{myexamplebox}

\subsection{Evaluation Metrics.}

To compare two annotated spans, we %
adapt the precision and recall of \citet{martinoFineGrainedAnalysisPropaganda2019} to alternatives.
Given two spans, $p$ with its label $l_p$, and $g$ with its set of labels $l_g$, respectively, 
and a normalizing constant $h$, these metrics compute a comparison score as follows:
\[ C(p, l_p, g, l_g, h) = \frac{|p \cap g|}{h} \times \delta(l_p, l_g) \]
\noindent $\delta$ is a similarity function. We use 
$\delta(x,y)=[x \in y ]$, 
where $[\cdot]$ is the Iverson bracket.

Let $G$ be the gold standard, and let $P$ be the prediction of a user or a system.
The precision for $P$ is computed by comparing each span in $P$ against all spans in $G$, and taking the score of the best-matching one:
\[\textit{Precision}(P, G)\!=\!\!\frac{\sum\limits_{(p, l_p)\in P}\!
\max\limits_{(g, l_g)\in G}\!C(p, l_p, g, l_g, |p|)}{|P|}\]
\noindent If there are no annotations in $P$ (i.e., $|P|=0$), we set precision to 1. 
This choice is inspired by the intuition that a loss in precision should result only from false predictions. If there are no such false predictions, then precision should not be harmed.

To calculate the recall, we exclude all the spans from the gold standard that contain a $\bot$. The rationale for this choice is that %
when a span has also been marked as ``no fallacy'', its annotation is considered optional. Therefore, we do not want to 
penalize models that do not provide an annotation for such a span. %
We define the set $G^-$, which is $G$ restricted to the spans that do not map to $\bot$, i.e., 
$G^-=\{(s,L) \in G \mid \bot \notin L\}$. %
The recall is then computed as:
\[\textit{Recall}(P, G)\!=\!\!\frac{\sum\limits_{(g, l_g)\in G^-}\!
\max\limits_{(p, l_p)\in P}\!C(p, l_p, g, l_g, |g|)}{|G^-|}\]
\noindent %
If  $|G^-|=0$, we set the recall to 1. 
The intuition is that a model should be penalized in recall only for the annotations it misses from the gold standard. If there are no such missed annotations, recall should not suffer. Appendix~\ref{app:metrics_edge_cases} shows how our metrics handle various edge cases.

The F1-score is computed as usual as the harmonic mean of precision and recall.
It is easy to see that our definitions of precision and recall fall back to the standard definitions if $G$ does not have alternatives and all spans comprise only a single sentence. 
In that case, the score $C$ is 1 if the spans are identical and identically labeled.  %
If there are no alternatives, our measures are also identical to the ones in %
\citet{martinoFineGrainedAnalysisPropaganda2019},
with one difference: we use the $\max$ instead of a sum in the definitions to select the best matching span. In this way, two neighboring spans with the same label do not achieve full precision or recall if the gold standard requires one contiguous span with that label. Using $\max$ instead of a sum also avoids the case where \citet{martinoFineGrainedAnalysisPropaganda2019}'s metric yields precision or recall scores exceeding one. This occurs when the gold standard contains overlapping spans with identical labels, affecting precision, or when the prediction includes such overlaps, affecting recall. %
However, their metric is equivalent to ours as long as (1) there are no alternatives in the gold standard, (2) 
neither the gold standard nor the prediction contains overlapping spans with the same label and (3) each span from the gold standard overlaps with at most one span with the same label from the predictions and vice versa. 
Examples, proofs, and further details are in Appendix~\ref{app:metrics}.

\section{MAFALDA Dataset}\label{sec:dataset}

\subsection{Source Datasets}\label{sec:dataset:sources}

We used four publicly available fallacy datasets to construct our benchmark:

We imported all 8,576 texts from \citet{sahaiBreakingInvisibleWall2021}. These are online discussions from Reddit, as in Example~(d) of Figure~\ref{fig:main}. %
We reconstructed the texts by concatenating the post of interest, the previous post (if available), and the title. The title was considered as a citation and was thus not annotated. This dataset contains sentences that were labeled as negative examples.

We imported all 336 texts from \citet{martinoFineGrainedAnalysisPropaganda2019}, which are from news outlets.
We imported all 583 texts from \citet{jinLogicalFallacyDetection2022}, which are either toy examples gathered from online quizzes (as in Example~(a)), or longer climate-related texts originating from news outlets.
Finally, we imported all 250 texts from \citet{goffredoFallaciousArgumentClassification2022}, which are American political debates (Example~(b)). 
We split these longer texts into shorter texts by concatenating the previous and following sentences of allegedly fallacious texts. 

\textbf{This gives us an English-language corpus of 9,745 texts, which is diverse in terms of linguistic terms and text length.} 
We removed URLs, emails, and phone numbers globally.

\subsection{Annotation}\label{sec:annotation:annotation}

The existing annotation schemes on our corpus varied a lot among papers: 
for example, only \citet{sahaiBreakingInvisibleWall2021} approached the annotation task as a binary classification, where annotators determine if a given text contains a specified type of fallacy.
The annotations process also varied greatly w.r.t. how consensus was obtained, as explained in Section~\ref{subsec:subjectivity}.

Therefore, we removed all annotations, and manually re-annotated, from scratch, 200 randomly selected texts from our merged corpus. Our sample mirrors the distribution of sources and the original labels in our corpus: it contains 124 texts from \citet{sahaiBreakingInvisibleWall2021}, 59 texts from \citet{jinLogicalFallacyDetection2022}, and 17 political debate texts from \cite{goffredoFallaciousArgumentClassification2022}.
We did not use the texts from \citet{martinoFineGrainedAnalysisPropaganda2019} we initially planned to use because they were more than 5,000 characters long.
Thus, annotating a single text would considerably bias the work towards Martino’s. However, the texts are part of our cleaned and homogenized dataset, and our goal is to include the annotations of these texts as we enlarge our manual annotation.

\textbf{LLMs were not involved} in the annotation process. We did \textbf{not involve crowd workers} either, because  
~33\%-46\% of Mturk workers are estimated to use ChatGPT~\cite{veselovsky2023artificial}. %
Hence, we annotated the texts ourselves.
Our task was (\textit{i}) identifying each argument in a text, (\textit{ii}) determining whether it is fallacious, (\textit{iii}) determining the span of the fallacy (as defined in Definition~\ref{def:span}), and (\textit{iv}) choosing the fallacy type(s).
We discussed each fallacious span together, and either converged on one annotation or permitted several alternative annotations for the same span. 
\textbf{We provide an explanation for each annotation}, and \textbf{we provide a completed template for each annotation}, as defined in Section~\ref{sec:tax}. 
For instance, 
Example (d) from Figure~\ref{fig:main}, which shows an \emph{appeal to ridicule}: the post argues against the possibility of working in finance with a law degree by exaggerating the position and thus portraying it as ridicule. Breaking down the example with the formal definition yields:
\begin{itemize}[noitemsep,nolistsep,leftmargin=4mm]
    \item $E_1$= The original poster
    \item $P$= It might be possible to work in finance with a law degree
    \item $E_2$= The author of the post.
    \item $P'$= Law school students are so intelligent that they can do any job, even surgeons.
\end{itemize}
Here, $E_i$ are entities (persons, organizations) or groups of entities, $P$ and $P'$ are premises, properties, or possibilities. %

The process took around 40 hours.
This corresponds to an average of 12 minutes per example, ranging from less than a minute for toy examples to half an hour when disagreement raised a debate. The total number of person-hours was 130.
Details about the annotators can be found in Appendix~\ref{app:annotators}.

\subsection{Statistics}

Our dataset comprises 9,745 texts, of which 200 texts have been annotated manually, with a total of 268 spans. 
Among these, 137 texts contained at least one span identified as fallacious,
while the remaining texts did not contain any fallacious spans.
The mean number of spans per text is 1.34. 

Among the 200 texts, 71 were initially labeled as non-fallacious.
However, our annotation found fallacies in some of these texts. This can be explained by the methodology from \citet{sahaiBreakingInvisibleWall2021}, where crowd workers check only one specific type of fallacy. If that fallacy is not present, the text is annotated as non-fallacious. Our annotation, however, spotted other fallacies in the text, and labeled them. 
In the end, we have 63 non-fallacious texts.

The dataset contains all the fallacies presented in Section~\ref{sec:tax}.
The three most frequent fallacies represent 1/4 of the dataset, while the least frequent fallacies appear less than three times. \textbf{71.5\% of the texts were annotated with a similar fallacy as the original one} (at least one fallacy of the source annotation was in the new annotation, or we agreed on a non-fallacious text). The difference is mainly because our taxonomy introduced new fallacies, such as \emph{appeal to ridicule}, and removed fallacies that we considered too vague or broad, such as \emph{intentional fallacy}. 
In some cases, we used a different granularity than in the source: while the source might say \emph{appeal to emotion}, we annotated, e.g., with \emph{appeal to fear}. We also permit several alternative annotations per span, which entails that the new annotations have, on average, more annotations per text than the source annotations (Original: 0.665, Ours: 1.34).

The dataset contains 203 spans, of which 65 
(i.e.,~28\%) contain at least two different (alternative) labels (see Example~\ref{ex:main}).
We computed the co-occurrence matrix of the fallacies.
Most fallacies do not co-occur too frequently (less than 30\% of the time) with another particular fallacy, which indicates that \textbf{our definitions of fallacies are broadly orthogonal}. 
However, there are two fallacies with high co-occurrence frequency: \appealtopity~ has a 100\% co-occurrence with \strawman~ and \appealtoworseproblem. However, this is because there is only one occurrence of \appealtopity~ in our dataset. The second one is \guiltbyassociation, which is 38\% of the time associated with \adhominem. This is explained by the fact that \guiltbyassociation~ and \adhominem{} are two types of the \emph{ad hominem} fallacy.
Complete statistics about our dataset are provided in Appendix~\ref{app:dataset}.

\section{Experiments}\label{sec:xp}
We will now evaluate the ability of state-of-the-art LLM to detect the fallacies in our benchmark. Our benchmark is not intended for training or fine-tuning, and hence, we study a zero-shot setting with basic prompts.
We are interested in the task of fallacy detection and classification of a given text, i.e.,~the input is a text, and the output is a list of annotated spans.

\subsection{Settings}
We study ChatGPT as well as 12 open-source models, covering different model sizes (Table~\ref{tab:result:partial}).
We use a bottom-up approach to evaluate our models starting at Level~2 granularity and extrapolate labels for Levels~1 and 0 based on these predictions, as our dataset includes three levels of granularity.

We employ a basic prompting approach that presents the model with our definition of a fallacy, the instruction to annotate the fallacies, the list of fallacies without their definitions, the corresponding text example, and the sentence to be labeled.
The detailed prompt can be found in Appendix~\ref{app:prompt}. Our experiments are conducted at the sentence level; spans are formed by grouping consecutive sentences with the same label.
A significant challenge with generative models is their inconsistent format output. Thus, we deem an output correct if it includes the name of the correct fallacy (or a part of it).
Details about the models can be found in Appendix~\ref{app:models}. 

\subsection{Results}
\label{sec:results}

\begin{table}[!ht]
    \centering
    \small

    \resizebox{\columnwidth }{!}{
     \begin{tabular}{lccc}
        \toprule
        & \multicolumn{3}{c}{MAFALDA} \\
        \cmidrule(lr){2-4} %
        Model & F1 Level 0~\textsuperscript{*} & F1 Level 1~\textsuperscript{*} & F1 Level 2\\
        \midrule
        Baseline random & 0.435 & 0.061 & 0.010  \\
        \midrule
        Falcon 7B& 0.397 & 0.130 & 0.022 \\
        LLAMA2 Chat 7B& 0.572 & 0.114 & 0.068 \\
        LLAMA2 7B& 0.492 & 0.148 & 0.038 \\
        Mistral Instruct 7B & 0.536 & 0.144 & 0.069 \\
        Mistral  7B & 0.450 & 0.127 & 0.044 \\
        Vicuna 7B & 0.494 & 0.134 & 0.051 \\
        WizardLM 7B & 0.490 & 0.087 & 0.036 \\
        Zephyr 7B & 0.524 & 0.192 & 0.098 \\
        LLaMA2 Chat 13B & 0.549 & 0.160 & 0.096 \\
        LLaMA2 13B & 0.458 & 0.129 & 0.039 \\
        Vicuna 13B & 0.557 & 0.173 & 0.100 \\
        WizardLM 13B & 0.520 & 0.177 & 0.093 \\
        \midrule
        GPT 3.5 175B & \textbf{0.627} & \textbf{0.201} & \textbf{0.138} \\
        \midrule
        Avg. Human & \textbf{0.749} & \textbf{0.352} & \textbf{0.186}\\
        on Sample & & & \\
        \bottomrule
    \end{tabular}
    }
    \begin{minipage}{\textwidth}
        \footnotesize
        \textsuperscript{*} Labels were extrapolated from Level 2.\\
    \end{minipage}
    \caption{Performance results of different models across different granularity levels in a zero-shot setting. Avg. human on sample concerns only the 20 subsamples of MAFALDA for the user study. Metrics are explained in Section~\ref{sec:metrics}.\label{tab:result:partial}}
    
\end{table}
    
Table \ref{tab:result:partial} shows the results across different granularity levels in a zero-shot setting, as evaluated using our metric (see Section~\ref{sec:metrics}). 
We added \textit{Baseline random}, a dummy model that predicts labels randomly following a uniform distribution.

At all levels of granularity, all models surpass the performance of the baseline model (except for Falcon on Level 0), indicating that they are successfully identifying certain patterns or features.
GPT~3.5 outperforms all other models at all levels. 
At Level 1, Zephyr 7B achieves comparable results to GPT 3.5, possibly thanks to the quality of its training dataset and/or engineering tricks, challenging the assumption that larger models are always more effective.
More surprisingly, LLaMA2 performs better in its 7B version than in its 13B version for Levels 0 and 1.
This phenomenon is in line with %
findings from \citet{DBLP:conf/emnlp/WeiKTL23}. 

We also investigate whether it makes a difference to prompt the models directly on Level 1 (as opposed to extrapolating Level~1 from Level~2). %
For Mistral Instruct and Zephyr, %
there is no significant difference: %
Mistral Instruct obtains an F1 score of 0.149, and Zephyr achieves an F1 score of 0.185.

\begin{table}[ht]
    \centering
    \small
    \resizebox{\columnwidth}{!}{
    \begin{tabular}{cccc}
    \toprule
     Gold Standard & F1 Level 0~\textsuperscript{*} & F1 Level 1~\textsuperscript{*} & F1 Level 2 \\
    \midrule
    User 1 & 0.616 & 0.310 & 0.119 \\
    User 2 & 0.649 & 0.304 & 0.098 \\
    User 3 & 0.696 & 0.253 & 0.093 \\
    User 4 & 0.649 & 0.277 & 0.144 \\
    \midrule
    MAFALDA & \textbf{0.749} & \textbf{0.352} & \textbf{0.186} \\
    \bottomrule
    \end{tabular}
    }
    \begin{minipage}{\textwidth}
        \footnotesize
        \textsuperscript{*} Labels were extrapolated from Level 2.\\
    \end{minipage}
    \caption{Cross-comparison of user annotations and the gold standard. Each annotation of the user study has been alternatively used as a gold standard to demonstrate the superiority of our own gold standard.}
    \label{tab:user_gs}
\end{table}

We also measure human performance on our dataset (which constitutes, to our knowledge, the first such study in the fallacy classification literature).
We aim to establish (\textit{i}) whether humans outperform language models for the task at hand and (\textit{ii}) whether humans agree more with our gold standard than among themselves. %
As human effort is more costly than running a language model (and even more so since we need engaged annotators who do not resort to ChatGPT or other LLMs),
we asked four other annotators to annotate 20 randomly chosen examples on the same task as the systems.
On these 20 samples, we compared the results of human annotators and LLMs.
The low scores of human annotators reported in Table~\ref{tab:user_gs} show that the task is difficult.
Still, \textbf{human participants outperform the language models} as shown in Table~\ref{tab:result:partial}: 
Contrary to what previous work has demonstrated~\cite{gilardiChatGPTOutperformsCrowdWorkers2023}, GPT-3.5 does not perform better than humans. 

Next, we study whether they agree more with our gold standard than among themselves.
We treat each annotator's work as a gold standard and assess the precision, recall, and F1 scores of the other annotators. Our gold standard achieves an F1 score of 0.186 on average for humans (see Table~\ref{tab:user_gs}), outperforming the best alternative, which scores 0.144. More details about the annotators and the results are in Appendix~\ref{app:annotators} and~\ref{app:results}, respectively. 

We analyze the errors in the two models, GPT-3.5 and Falcon, focusing on their performance at Levels 1 and 2, alongside user study annotations. Our first goal is to determine whether the models, specifically the highest-performing GPT-3.5 and the lowest-performing Falcon, exhibit controlled or uncontrolled behavior in their output generation. 
While both models can produce nonsensical outputs, Falcon often predicts multiple irrelevant fallacies and significantly more unknown labels than GPT-3.5.

Our second goal is to determine which fallacy type is the most challenging. The analysis reveals that humans and models struggle with the fallacies of appeal to emotion. We hypothesize that emotions often appear in texts without necessarily constituting a fallacy, which complicates the distinction between emotional texts and fallacious texts that use appeals to emotion.
More details about this analysis are in Appendix~\ref{app:error_analysis}.

\section{Conclusion}\label{sec:conclusion}
We have presented MAFALDA, a unified dataset designed for fallacy detection and classification. This dataset integrates four pre-existing datasets into a cohesive whole, achieved through developing a new, comprehensive taxonomy. This taxonomy aligns %
publicly available taxonomies dedicated to fallacy detection. 
We manually annotated 200 texts from our dataset and provided an explanation in the form of a completed template for each of them. 
The disjunctive annotation scheme we proposed embraces the subjectivity of the task and allows for several alternative annotations for the same span. 
We have further demonstrated the capabilities of various large language models in zero-shot fallacy detection and classification at the span level.
While Level 0 classification shows good results, Levels 1 and 2 are largely out of reach of LLMs in zero-shot settings.
We hope that our benchmark will enable researchers to improve the results of this challenging task.

Future work includes expanding into few-shot settings and exploring advanced prompting techniques, such as chain-of-thought, using the template-based definitions of fallacy and the taxonomy we provided. Furthermore, we believe that using a top-down approach, i.e.,~from Level 0 to Level 2 of our taxonomy, may provide better results than the bottom-up approach we used in our experiments. 
Regarding our disjunctive annotation scheme, we are interested in exploring its use in other NLP domains.
Lastly, we plan to enrich the dataset with more annotated examples for model fine-tuning.\\

\textbf{Acknowledgement.} This work was partially funded by the NoRDF project (ANR-20-CHIA-0012-01), the SINNet project (ANR-23-CE23-0033-01) and Amundi Technology.

\section*{Limitations}\label{sec:limitations}

Our work provides a dataset that may contain sensitive content such as racism, apologies for committing crimes, and misogyny. We believe that this is unavoidable in our fight against fake news and manipulative content. 

One limitation of our benchmark is that we may have been biased while annotating it. Our collective bias is limited because we come from diverse cultural backgrounds, countries, religions, and political convictions, but it may still exist. 
We have also made efforts to mitigate bias during our annotation, with clear guidelines, or by achieving a consensus or at least providing strong arguments in the form of a completed template as described in the formal definitions of Appendix~\ref{app:fallacy_list}. 

Our dataset has the potential for misuse in training systems that could be exploited for manipulative purposes, such as crafting more convincing fallacious arguments or disinformation campaigns.
Finally, models trained on this problem may wrongly label a text as fallacious. They must thus not be used to flag a text as fallacious without manual verification. %

A further consideration is the size of our dataset. It is relatively small due to the time-intensive nature of the annotation process. It is thus not suited for fine-tuning, but rather intended for evaluating large language models in zero-shot and few-shot settings.  %

\bibliography{references}

\appendix

\section{Definitions of the fallacies}\label{app:informal_def}\label{app:fallacy_list}

In the following, we provide, for each fallacy, its informal definition, its formal definition, and a toy example. 
We start by describing the variables/placeholders used in the formal templates.
\begin{itemize}
    \item $A =$ attack
    \item $E =$ entity (persons, organizations) or group of entities
    \item $P, P_i =$ premises, properties, or possibilities
    \item $C =$ conclusion
\end{itemize}
\noindent The following definitions are inspired by \citet{bennettLogicallyFallaciousUltimate2012} and have been adapted to be more generic.

\subsection{Fallacies of Emotion}

\fallacy{Appeal to Anger}{This fallacy involves using anger or indignation as the main justification for an argument, rather than logical reasoning or evidence.}{$E$ claims $P$. $E$ is outraged. Therefore, $P$. Or $E_1$ claims $P$. $E_2$ is outraged by $P$. Therefore, $P$ (or $\neg P$ depending on the situation).}{The victim’s family has been torn apart by this act of terror. Put yourselves in their terrible situation, you will see that he is guilty.}{$E$ (the speaker) claims $P$ (the accused is guilty) and expresses outrage. Therefore, $P$ (guilt).}

\fallacy{Appeal to Fear}{This fallacy occurs when fear or threats are used as the main justification for an argument, rather than logical reasoning or evidence.}{If $\neg P_1$, something terrible $P_2$ will happen. Therefore, $P_1$.}{If you don't support this politician, our country will be in ruins, so you must support them.}{If $\neg P_1$ (not supporting the politician), then $P_2$ (country in ruins) will happen. Therefore, $P_1$ (must support the politician).}

\fallacy{Appeal to Pity}{This fallacy involves using sympathy or compassion as the main justification for an argument, rather than logical reasoning or evidence.}{$P$ which is pitiful, therefore $C$, with only a superficial link between $P$ and $C$}{He's really struggling, so he should get the job despite lacking qualifications.}{$P$ (he’s struggling) is presented as a pitiful situation, leading to $C$ (he should get the job), despite a merely superficial link between $P$ and $C$.}

\fallacy{Appeal to Positive Emotion}{This fallacy occurs when a positive emotion -- like hope, optimism, happiness, or pleasure -- is used as the main justification for an argument, rather than logical reasoning or evidence.}{$P$ is positive. Therefore, $P$.}{Smoking a cigarette will make you look cool, you should try it!}{$P$ (smoking cigarettes looks cool) leads to $P$ (try smoking).}

\fallacy{Appeal to Ridicule}{This fallacy occurs when an opponent's argument is portrayed as absurd or ridiculous with the intention of discrediting it.}{$E_1$ claims $P$. $E_2$ makes $P$ look ridiculous, by misrepresenting $P$ ($P$'). Therefore, $\neg P$.}{There's a proposal to reduce carbon emissions by 50\% in the next decade. What’s next? Are we all going to stop breathing to reduce CO2?}{$E_1$ (unspecified entity) claims $P$ (proposal to reduce the carbon emissions). $E_2$ (the speaker) makes $P$ looks ridiculous by suggesting an extreme scenario $P'$ (stop breathing). Therefore, $\neg P$ (reducing carbon emissions is unreasonable).}

\fallacy{Appeal to Worse Problems}{This fallacy involves dismissing an issue or problem by claiming that there are more important issues to deal with, instead of addressing the argument at hand. This fallacy is also known as the "relative privation" fallacy.}{$P_1$ is presented. $P_2$ is presented as a best-case. Therefore, $P_1$ is not that good. OR $P_1$ is presented. $P_2$ is presented as a worst-case. Therefore, $P_1$ is very good.}{Why worry about littering when there are bigger problems like global warming?}{$P_1$ (littering) is compared to $P_2$ (global warming), which is a worse problem, leading to the conclusion that $P_1$ is not important.}

\subsection{Fallacies of Logic}

\fallacy{Causal Oversimplification}{This fallacy occurs when a complex issue is reduced to a single cause and effect, oversimplifying the actual relationships between events or factors.}{$P_1$ caused $C$ (although $P_2$, $P_3$, $P_4$, etc.\ also contributed to $C$.)}{There is an economic crisis in the country, the one to blame is the president.}{$P_1$ (the president) caused $C$ (economic crisis),  while ignoring other contributing factors ($P_2$ (worldwide economical context), $P_3$ (previous policies), etc.).}

\fallacy{Circular Reasoning}{This fallacy occurs when an argument assumes the very thing it is trying to prove, resulting in a circular and logically invalid argument.}{$C$ because of $P$. $P$ because of $C$. OR $C$ because $C$.}{The best smartphone is the iPhone because Apple creates the best products.}{$C$ (iPhone is the best smartphone) because $P$ (Apple creates the best products), which in turn is justified by the claim $C$.}

\fallacy{Equivocation}{This fallacy involves using ambiguous language or changing the meaning of a term within an argument, leading to confusion and false conclusions.}{No logical form: $P_1$ uses a term $T$ that has a meaning $M_1$. $P_2$ uses the term $T$ with the meaning $M_2$ to mislead.}{The government admitted that many cases of credible UFOs (Unidentified flying objects) have been reported. Therefore, that means that Aliens have already visited Earth.}{$P_1$ (many cases of credible UFOs have been reported) uses the term UFO with the meaning $M_1$ (unidentified flying objects). $P_2$ (aliens have already visited Earth) uses UFO with a different meaning $M_2$ (implying that aliens = UFOs), misleading the conclusion.}

\fallacy{Fallacy of Division}{This fallacy involves assuming that if something is true for a whole, it must also be true of all or some of its parts.}{$E_1$ is part of $E$, $E$ has property $P$. Therefore, $E_1$ has property $P$.}{The team is great, so every player on the team must be great.}{$E_1$ (every player) is part of $E$ (the team). $E$ has the property $P$ (great),  then $E_1$ also has $P$.}

\fallacy{False Analogy}{This fallacy involves making an analogy between two elements based on superficial resemblance.}{$E_1$ is like $E_2$. $E_2$ has property $P$. Therefore, $E_1$ has property $P$. (but $E_1$ really is not too much like $E_2$)}{We should not invest in Space Exploration. It's like saying that a person in debt should pay for fancy vacations.}{$E_1$ (a country in debt plans to explore space) is linked to $E_2$ (a family in debt plans fancy vacations). $E_2$ has property $P$ (expensive and not advisable), implying $E_1$ should also have $P$.}

\fallacy{False Causality}{This fallacy involves incorrectly assuming that one event causes another, usually based on temporal order or correlation rather than a proven causal relationship.}{$P$ is associated with $C$ (when the link is mostly temporal and not logical). Therefore, $P$ causes $C$.}{After the rooster crows, the sun rises; therefore, the rooster causes the sunrise.}{$P$ (rooster crows) is associated with $C$ (sunrise), but the link is temporal, not causal, leading to the false conclusion that $P$ causes $C$.}

\fallacy{False Dilemma}{This fallacy occurs when only two options are presented in an argument, even though more options may exist.}{Either $P_1$ or $P_2$, while there are other possibilities. OR Either $P_1$, $P_2$, or $P_3$, while there are other possibilities.}{You're either with us, or against us.}{Presents a choice between $P_1$ (with us) and $P_2$ (against us), excluding other possibilities.}

\fallacy{Hasty Generalization}{This fallacy occurs when a conclusion is drawn based on insufficient or unrepresentative evidence.}{Sample $E_1$ is taken from population $E$. (Sample $E_1$ is a very small part of population $E$.) Conclusion $C$ is drawn from sample $E_1$.}{I met two aggressive dogs, so all dogs must be aggressive.}{A small sample $E_1$ (two aggressive dogs) is taken from a larger population $E$ (all dogs). Therefore $C$ (all dogs are aggressive).}

\fallacy{Slippery Slope}{This fallacy occurs when it is claimed that a small step will inevitably lead to a chain of events, resulting in a significant negative outcome.}{$P_1$ implies $P_2$, then $P_2$ implies $P_3$,... then $C$ which is negative. Therefore, $\neg P_1$.}{If we allow kids to play video games, they will see fights, guns, and violence, and then they'll become violent adults.}{$P_1$ (allowing kids to play video games) implies $P_2$ (seeing fights, guns, and violence), which in turns implies $P_3$ (to like violence, etc.) leading to $C$ (kids becomes violent adults). Therefore, $\neg P_1$.}

\fallacy{Strawman Fallacy}{This fallacy involves misrepresenting an opponent's argument, making it easier to attack and discredit.}{$E_1$ claims $P$. $E_2$ restates $E_1$’s claim (in a distorted way $P'$). $E_2$ attacks ($A$) $P'$. Therefore, $\neg P$.}{He says we need better internet security, but I think his panic about hackers is overblown.}{$E_1$ (an unspecified person (He)) claims $P$ (need for better internet security), $E_2$ (the speaker)  distorts the claim as $P'$ (panic about hackers). Therefore $\neg P$.}

\subsection{Fallacies of Credibility}

\fallacy{Abusive Ad Hominem}{This fallacy involves attacking a person's character or motives instead of addressing the substance of their argument.}{$E$ claims $P$.  $E$'s character is attacked ($A$). Therefore, $\neg P$.}{``John says the earth is round, but he's a convicted criminal, so he must be wrong.''}{$E$ (John) claims $P$ (the earth is round). John's character is attacked $(A)$ (being a criminal). Therefore,  $\neg P$ (the earth is not round).}

\fallacy{Ad Populum}{This fallacy involves claiming that an idea or action is valid because it is popular or widely accepted.}{A lot of people believe/do $P$. Therefore, $P$. OR Only a few people believe/do $P$. Therefore, $\neg P$.}{Millions of people believe in astrology, so it must be true.}{Many people believe in  $P$ (astrology). Therefore, $P$ (astrology is true).}

\fallacy{Appeal to Authority}{This fallacy occurs when an argument relies on the opinion or endorsement of an authority figure who may not have relevant expertise or whose expertise is questionable. When applicable, a scientific consensus is not an appeal to authority.}{$E$ claims $P$ (when $E$ is seen as an authority on the facts relevant to $P$). Therefore, $P$.}{A famous actor says this health supplement works, so it must be effective.}{$E$ (famous actor) claims $P$ (the health supplement works). Therefore,  $P$ (it must be effective).}

\fallacy{Appeal to Nature}{This fallacy occurs when something is assumed to be good or desirable simply because it is natural, while its unnatural counterpart is assumed to be bad or undesirable.}{$P_1$ is natural. $P_2$ is not natural. Therefore, $P_1$ is better than $P_2$. OR $P_1$ is natural, therefore $P_1$ is good.}{Herbs are natural, so they are better than synthetic medicines.}{$P_1$ (herbs are natural) and $P_2$ (synthetic medicines are not natural), leading to $P_1$ is better than $P_2$.}

\fallacy{Appeal to Tradition}{This fallacy involves arguing that something should continue to be done a certain way because it has always been done that way, rather than evaluating its merits.}{We have been doing $P$ for generations. Therefore, we should keep doing $P$. OR Our ancestors thought $P$. Therefore, $P$.}{We've always had a meat dish at Thanksgiving, so we should not change it.}{$P$ (always had a meat dish at Thanksgiving) should continue. Therefore, continue $P$.}

\fallacy{Guilt by Association}{This fallacy involves discrediting an idea or person based on their association with another person, group, or idea that is viewed negatively.}{$E_1$ claims $P$. Also $E_2$ claims $P$, and $E_2$'s character is attacked ($A$). Therefore, $\neg P$. OR $E_1$ claims $P$. $E_2$'s character is attacked ($A$) and is similar to $E_1$. Therefore $\neg P$.}{Alice believes in climate change, just like the discredited scientist Bob, so her belief must be false.}{$E_1$ (Alice) claims $P$ (belief in climate change). $E_2$ (Bob) also claims $P$. However $E_2$'s character $(A)$ is attacked (being discredited). Therefore $\neg P$.}

\fallacy{Tu Quoque}{This fallacy occurs when someone's argument is dismissed because they are accused of acting inconsistently with their claim, rather than addressing the argument itself.}{$E$ claims $P$, but $E$ is acting as if $\neg P$. Therefore $\neg P$.}{Laura advocates for healthy eating but was seen eating a burger, so her advice on diet is invalid.}{$E$ (Laura) claims $P$ (advocates for healthy eating), but $E$ is acting as if $\neg P$ (eating a burger, which is unhealthy eating). Therefore $\neg P$ (advice on diet is invalid).}

Our categorization of fallacies into logic, emotion, and credibility is based on the primary aspect of the fallacy that leads to an invalid or weak argument. 
In practice, some fallacies %
could be argued to fit into more than one category. %

\section{Comparison of Fallacy Types}\label{app:comparison_of_types}

Previous works have studied a large number of different fallacy types. The earliest works focused on \emph{ad hominem}, while later works included dozens of other types. To build our taxonomy, we tried to unify most fallacy types in the literature. Table~\ref{tab:fallacies_per_papers} shows each type of fallacy studied by each paper that proposed a dataset. Most fallacies from our taxonomy are part of at least two already existing datasets. Based on our definition (Section~\ref{sec:def}), rhetorical techniques that are not based on an actual argument are not considered fallacies. Thus, we did not include techniques such as \emph{repetition} or \emph{slogans}. During the initial annotation phase, we observed that the \emph{red herring} fallacy was too vague, so we replaced it with more precise sub-categories, such as \appealtoworseproblem. This explains why \appealtoworseproblem~, which is present in only one other dataset, is part of our taxonomy. Similarly, during the annotation, we found multiple examples of \emph{fallacy of division}, which is related to \emph{hasty generalization} but does not fit its description. Hence, we added \emph{fallacy of division} in the taxonomy.

\begin{sidewaystable*}
\tiny
\begin{tabular}{lp{9em}|p{5em}|p{6em}|p{2cm}|l|p{2cm}|p{1.5cm}|p{3em}p{4em}|p{5em}|p{5em}|l}

\cline{5-8}

\multicolumn{2}{c}{Our Taxonomy} & & & \multicolumn{4}{|c|}{\citet{alhindiMultitask2022}}  & & & & & \\

\cline{5-8}
\cline{5-8}

Level 1 & Level 2 &  \citet{habernalNameCallingDynamicsTriggers2018} & \citet{delobelleComputationalAdHominem2019} & \citet{martinoFineGrainedAnalysisPropaganda2019} \cite{balalauStageAudiencePropaganda2021} & \citet{jinLogicalFallacyDetection2022}  & \citet{musiCOVID2022} & \citet{habernalArgotario2017} & \multicolumn{2}{c}{\citet{goffredoFallaciousArgumentClassification2022}} & \citet{reisertRiposte2019} & \citet{sahaiBreakingInvisibleWall2021} & \citet{hong2023closer} \\

\hline

\rowcolor{green!25}  & & & & & Ad Hominem & & &  & General & & & \cellcolor{gray!25} \\
\rowcolor{green!25} & Abusive Ad Hominem & abusive & abusive attack & & & & & & & & & \cellcolor{gray!25} \\
\rowcolor{green!25} & Tu Quoque & tu quoque & & whataboutism & & & & & Tu quoque & & & \cellcolor{gray!25} \\
\rowcolor{green!25} & & circumstantial & & & & & & & & & & \cellcolor{gray!25} \\
\rowcolor{green!25} & & bias & & & & & & & Bias ad hominem & & & \cellcolor{gray!25} \\
\rowcolor{green!25} & Guilt by Association & guilt by association & & reductio ad hitlerum & & & & & & & & \cellcolor{gray!25} \\
\rowcolor{green!25} & & & name calling & name calling & & & & & Name-calling & & & \cellcolor{gray!25} \\
\rowcolor{green!25} & & & & doubt & & & \multirow{-12}{*}{ad hominem} & \multirow{-12}{4em}{ad hominem} & & & & \cellcolor{gray!25} \\

\cline{2-12}

\rowcolor{green!25} & Ad Populum & & & bandwagon & Ad Populum & & &  & Popular opinion & & Appeal to Majority & \cellcolor{gray!25} \\
\rowcolor{green!25} & Appeal to Nature & & & & & & & & & & Appeal to Nature & \cellcolor{gray!25} \\

\cline{9-9}

\rowcolor{green!25} & Appeal to Tradition & & & & & & & & & & Appeal to Tradition & \cellcolor{gray!25} \\

\cline{2-8} \cline{10-12}

\rowcolor{green!25} & & & & & Fallacy of Credibility & Appeal to Inappropriate Authority & irrelevant authority & & False authority & & Appeal to Authority & \cellcolor{gray!25} \\
\rowcolor{green!25} \multirow{-20}{4em}{Fallacy of Credibility}  & \multirow{-3}{10em}{Appeal to False Authority} & & & \multirow{-3}{10em}{appeal to false authority} & & & & \multirow{-5}{4em}{Appeal to authority} & without evidence & & & \cellcolor{gray!25} \\

\cline{1-12}

\rowcolor{orange!25}  & & & & & & Evading the Burden of Proof & & & & & & \cellcolor{gray!25} \\
\rowcolor{orange!25} & Causal Oversimplification & & & causal oversimplification & & & & & & & & \cellcolor{gray!25} \\
\rowcolor{orange!25} & & & & & & \multirow{-2}{8em}{Cherry Picking of Evidence} & & & & &  & \cellcolor{gray!25} \\
\rowcolor{orange!25} & \multirow{-2}{*}{Hasty Generalization} & & & & \multirow{-2}{*}{Faulty Generalization} & Hasty Generalization & Hasty Generalization & & & \multirow{-2}{3em}{Hasty Generalization} &  \multirow{-2}{5em}{ Hasty Generalization } & \cellcolor{gray!25} \\
\rowcolor{orange!25} & & & & & & False Cause & & & &  & & \cellcolor{gray!25} \\
\rowcolor{orange!25} & \multirow{-2}{*}{False Causality} & & & & \multirow{-2}{*}{False Causality} & Post Hoc (Correlation presented as Causation) & & \multirow{-2}{4em}{False cause} & & \multirow{-2}{3em}{Questionable Cause} & & \cellcolor{gray!25} \\
\rowcolor{orange!25} & Circular Reasoning & & & & Circular Claim & & & & & Begging the Question & & \cellcolor{gray!25} \\
\rowcolor{orange!25} & False Dilemma & & & black-and-white fallacy & False Dilemma & & & & & & Black-or-White & \cellcolor{gray!25} \\
\rowcolor{orange!25} & Slippery Slope & & & & & & & & Slippery Slope & & Slippery Slope & \cellcolor{gray!25} \\
\rowcolor{orange!25} & False Analogy & & & & & False Analogy & & & & & & \cellcolor{gray!25} \\
\rowcolor{orange!25} & Straw Man & & & straw man & Fallacy of Extension & Strawman & & & & & & \cellcolor{gray!25} \\
\rowcolor{orange!25} & Fallacy of Division & & & & & & & & & & & \cellcolor{gray!25} \\
\rowcolor{orange!25}& & & & & Deductive Fallacy & & & & & & & \cellcolor{gray!25} \\
\rowcolor{orange!25} & & & & red herring & Fallacy of Relevance & Red Herring & Red Herring & & & Red Herring & & \cellcolor{gray!25} \\
\rowcolor{orange!25} \multirow{-20}{3em}{Fallacy of Logic} & Equivocation & \cellcolor{gray!25} & & obfusc. int. vagueness confusion & Equivocation & Vagueness & & & & & & \cellcolor{gray!25} \\

\cline{1-2} \cline{4-12}
 
\rowcolor{gray!25} \cellcolor{blue!15} & \cellcolor{blue!15}  & & & thought-terminating cliches & & & & & & & & \cellcolor{gray!25} \\
\rowcolor{gray!25} \cellcolor{blue!15} & \cellcolor{blue!15}  & & & exaggeration/minimization & & & & & & & & \cellcolor{gray!25} \\
\rowcolor{gray!25} \cellcolor{blue!15} & \cellcolor{blue!15}  & & & repetition & & & & & & & & \cellcolor{gray!25} \\
\rowcolor{gray!25} \cellcolor{blue!15} & \cellcolor{blue!15}  & & & slogans & & & & Slogan & & & & \cellcolor{gray!25} \\
\rowcolor{gray!25} \cellcolor{blue!15} & \cellcolor{blue!15} & \multirow{-8}{4em}{Technique based on use of vocabulary} & & loaded language & & & & & Loaded Language & & & \cellcolor{gray!25} \\
\rowcolor{blue!15} & Appeal to Positive Emotion & & & flag-waving & & & & & Flag waving & & & \cellcolor{gray!25} \\
\rowcolor{blue!15}& Appeal to Anger & & & & & & & & & & & \cellcolor{gray!25} \\
\rowcolor{blue!15}& Appeal to Fear & & & appeal to fear/prejudice & & & & & Appeal to fear & & & \cellcolor{gray!25} \\
\rowcolor{blue!15}& Appeal to Pity & & & & & & & & Appeal to pity & & & \cellcolor{gray!25} \\
\rowcolor{blue!15}& Appeal to Ridicule & & & & \multirow{-6}{*}{Appeal to Emotion} & & \multirow{-6}{4em}{Appeal to Emotion } & \multirow{-6}{4em}{Appeal to Emotion } & & & & \cellcolor{gray!25} \\
\rowcolor{blue!15}& Appeal to Worse Problem & & & & & & & & & & Appeal to Worse Problems & \cellcolor{gray!25} \\
\rowcolor{blue!15}& & & & & Intentional & & & & & & & \cellcolor{gray!25} \\
\rowcolor{blue!15}\multirow{-18}{3em}{Fallacy of Emotion} & & & & red herring & Fallacy of Relevance & Red Herring & & & & Red Herring & & \multirow{-60}{5em}{\cellcolor{gray!25}Probability Proposition Quantification Syllogism Ambiguity Inconsistency Irrelevance Insufficiency Inappropriate Presumption}\\

\hline
\end{tabular} 
\caption{List of fallacies per paper and in our taxonomy.}\label{tab:fallacies_per_papers}
\end{sidewaystable*}

\section{Additional Examples}\label{app:additional_examples}

Example~\ref{ex:political} is an \emph{appeal to positive emotion}. Using our formal definition, it has the following key component:
\begin{itemize}
    \item $P$= pride in the military's strength and state of preparedness
\end{itemize}
\noindent Example~\ref{ex:main} illustrates a case of alternative labels, where the text can be a \emph{causal oversimplification} or a \emph{false causality}. In the context of \emph{causal oversimplification}, the scenario can be deconstructed according to the formal definition as follows:
\begin{itemize}
    \item $P_1$= Winner of the last primary election
    \item $C$= He/she will win the general election
    \item $P_1, P_2,\ldots$: Factors including political dynamics, the opponent in the race, etc.
\end{itemize}
As a \emph{false causality}, the structure is:
\begin{itemize}
    \item $P$= Winner of the last primary election
    \item $C$= He/she will win the general election
\end{itemize}

\section{Annotation Guidelines for Identifying Fallacious Arguments}\label{sec:guideline}

The task of annotating a text with fallacies can be decomposed into several steps: First, determine if the text contains an argument and what the premises and conclusion are. Then, the span must be delimited. Finally, an adequate label must be chosen. 
For the construction of our gold standard, annotators used %
Doccano\footnote{\url{https://github.com/doccano/doccano}}, and  followed these guidelines: 

\begin{enumerate}
    \item \textbf{Consensus Requirement:} Before finalizing annotations for any given text, annotators should try to reach a consensus. This collaborative approach ensures consistency and accuracy in the identification of fallacious arguments. In instances where consensus is unattainable, the differing viewpoints regarding potential fallacies should be noted as alternative interpretations, as detailed in Section~\ref{sec:annotation}.

    \item \textbf{Resource Utilization:} Annotators are encouraged to consult various resources, including Google Search, Wikipedia, and books on argumentation. However, using Large Language Models, such as ChatGPT, is prohibited to prevent potential bias or contamination in the annotations.

    \item \textbf{Reference Material:} For definitions and clarifications:
    \begin{itemize}
        \item Refer to the definitions of an argument and a fallacy as outlined in Section \ref{sec:def} and Appendix~\ref{sec:edge}.
        \item Consult the Appendix \ref{app:informal_def} for detailed formal and informal definitions of individual fallacies.
        \item Follow the definition of spans detailed in Section \ref{sec:annotation}
    \end{itemize}

    \item \textbf{Annotation Protocol:}
    \begin{itemize}
        \item Upon reaching a consensus, annotators must document their rationale, aligning their reasoning with the formal definitions provided.
    \item Annotators are encouraged to add useful comments. This includes identifying text segments that may require special post-processing or additional review. %
    \end{itemize}
\end{enumerate}

\noindent Our annotation guidelines are also available on the Web page of our project, \url{https://github.com/ChadiHelwe/MAFALDA/}. %

\subsection{Edge Cases}\label{sec:edge}

We provide additional information to help the annotators with edge cases.

In our definition (as in \citet{fallacy_def}), a fallacy is always an argument in the sense of Definition~\ref{def:arg}, i.e.,~\textbf{a fallacy is always of the form ``\emph{A, B, C}, ... therefore \emph{X}'' or of the form ``\emph{X} because \emph{A, B, C}, ...'', or it can be rephrased into these forms}.
Hence, false assertions are not, \textit{per se}, fallacies. 
For example, ``\textit{Paris is the capital of England}'' is a false claim. However, it is not a fallacy because it is not an argument. 
The same goes for generalizations: ``\textit{All Americans love Trump}'' is false, but not a fallacy. 
An insult (such as ``\textit{You are too stupid}''), likewise, is not a fallacy\footnote{It becomes a fallacy when it is used as the premise of an argument, as in ``\textit{You are stupid, therefore what you say can't be true.}''}. Slogans (such as ``\textit{America first!}''), likewise, are not fallacies in our definition, even if other works classify them as propaganda~\cite{martinoFineGrainedAnalysisPropaganda2019}. %
An \emph{appeal to emotion} (such as ``\textit{Think of the poor children!}''), likewise, is not a fallacy by itself. 
It becomes a fallacy only when used as the premise of a fallacious argument, as in: ``\textit{Think of the poor children, and [therefore] vote for me!''}.  
But not every argument that appeals to emotion is automatically fallacious. 
For instance, the argument ``\textit{During a Covid-19 pandemic, you should wear a mask in public transport because otherwise you could get infected}'' appeals to fear. However, it is still a valid argument because the premise does entail the conclusion. 
Even if the premises of an argument are factually false, the argument is not necessarily fallacious. For example, ``\textit{All Americans love Trump, and therefore Biden loves Trump}'' is an argument that rests on a false premise -- but it is not fallacious because the premise indeed entails the conclusion in the sense of \cite{tina}: if the premise were true, the conclusion would be true as well. 
\textbf{The fallaciousness of an argument is thus largely independent of the truth values of its components.} 

Finally, \textbf{the description of fallacious reasoning is not automatically fallacious.} 
For example, ``\textit{You should wear a tin foil hat because it protects you against mind control}'' is a fallacy (because the tin foil hat does not protect against mind control of any known form). 
However, the following is a factual assertion, not a fallacy: ``\textit{Some people wear tin foil hats because they are afraid of mind control}''.

\section{Annotators}\label{app:user_study}
\label{app:annotators}

\subsection{Annotators}

We conducted two annotation phases, one for the annotation of the gold labels, which resulted in the MAFALDA dataset, and one for the user study. Here is a description of the background of the annotators.

\subsubsection{Gold Standard Annotators}\label{app:annotators:gold}

The gold standard was produced by the authors of the paper, who have the following characteristics:
\begin{itemize}
    \item Nationality: Lebanese, Gender: Male, Native language: Arabic, Education: Master's degree, Occupation: Ph.D. student in computer science.

    \item Nationality: French, Gender: Male, Native language: French, Education: Master's degree, Occupation: Ph.D. student in computer science.

    \item Nationality: French, Gender: Male, Native language: French, Education: Ph.D. degree, Occupation: Post-doctoral researcher in computer science.

    \item Nationality: French, Gender: Female, Native language: French, Education: Ph.D. degree, Occupation: Professor in computer science.

    \item Nationality: German,  Gender: Male, Native language: German, Education: Ph.D. degree, Occupation: Professor in computer science.
\end{itemize}

\textbf{Compensation}: The annotators are the paper's authors and did not receive compensation for the annotations.

\textbf{Biases and Limitations}: The annotators are all authors of the paper. They are working in the computer science field

\subsubsection{User Study Annotators}\label{app:annotator:study}

The user study annotations were provided by the following 4 persons:
\begin{itemize}
    \item Nationality: Lebanese, Gender: Male, Native Language: Arabic%
    Education: Master's degree in mechanical engineering, Occupation: Statistics Expert.%

    \item Nationality: French, Gender: Male, Native Language: French, %
    Education: Master's degree in big data and data science, Occupation: Ph.D. Student in computer science.

    \item Nationality: Morrocan, Gender: Female, Native Language: French, %
    Education: Ph.D. degree, Occupation: Data scientist.
 
    \item Nationality: French, Gender: Male, Native Language: French, %
    Education: Master's degree in machine learning, Occupation: Ph.D. Student in computer science.
\end{itemize}

\textbf{Compensation}: The annotators were volunteers and were not compensated for the annotations.

\subsection{Insights from the User Study Annotators}\label{app:annotators_insights}

The annotation process was very time-consuming, with some annotators taking up to four hours to complete their task for the 20 examples. 
One annotator humorously questioned their normality, stating, ``\textit{I don’t know if I’m a normal human, but I found it difficult! :)}'' while another jokingly expressed regret over accepting the task. These comments reflect the general sentiment about the task's complexity. The annotators often struggled with specific examples, such as \textit{``Reasonable regulations don’t lead to the fed keeping lists and someday coming after all gun owners to suppress the working class''}, which has been annotated differently by each user such as an \emph{ad populum} and \emph{false causality} fallacy 
while it is not a fallacy. 
This is often due to over-complicated sentences.

\section{Dataset}\label{app:dataset}

\begin{table}[ht]
\centering
\resizebox{\columnwidth}{!}{%
\begin{tabular}{lrr}
\toprule
         Source Dataset & Non-annotated & Annotated \\
        \midrule
        \citet{sahaiBreakingInvisibleWall2021} & 640 (7,812) & 71 (53) \\
        \citet{jinLogicalFallacyDetection2022} & 524 & 59 \\
        \citet{martinoFineGrainedAnalysisPropaganda2019} & 336 & 0 \\
        \citet{goffredoFallaciousArgumentClassification2022} & 233 & 17 \\
        \bottomrule
        TOTAL & 1733 (7,812) & 137 (53) \\
\end{tabular}%
}
\caption{Distribution of text from the initial source and from the final re-annotated dataset. Numbers in parenthesis are for non-fallacious texts.}\label{tab:distribution_origin_sample}

\end{table}

\begin{table}[ht]
\centering
\resizebox{\columnwidth}{!}{%
\begin{tabular}{lrrrrrrr}
\toprule
Number of Spans & 0 & 1 & 2 & 3 & 4 & 5 & 6 \\
\midrule
w/ counting alternatives & 63 & 70 & 32 & 18 & 8 & 6 & 3 \\
w/o counting alternatives & 63 & 95 & 23 & 15 & 3 & 1 & 0 \\
\bottomrule
\end{tabular}%
}
\caption{Number of text with N spans. The first line considers alternatives, i.e.,~a disjunction of two labels for a span will count as two annotations. Conversely, in the second line, an alternative will count as one annotation. This allows for comparing the usage of alternatives in our annotations.}
\label{tab:number_span}

\end{table}

\begin{figure*}[!hbp]
    \centering
    \includegraphics[width=\textwidth]{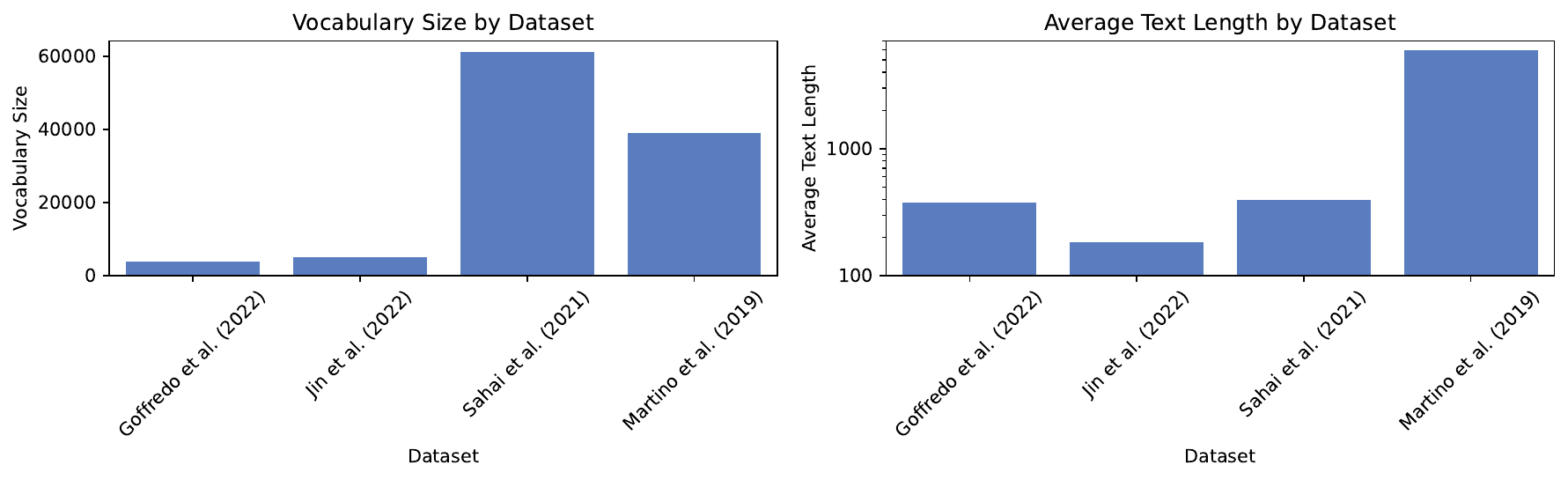}
    \caption{Statistics about our source datasets. The left graphic shows the vocabulary size, while the right graphic shows the average length of the texts. }%
    \label{fig:source_dataset}
\end{figure*}

Table~\ref{tab:distribution_origin_sample} presents statistics about our dataset: the source of each text, Table~\ref{tab:number_span} displays the number of annotations in the 200 texts, and Table~\ref{tab:number_fallacy} presents the frequency of each fallacy.

The left-hand side of Figure~\ref{fig:source_dataset} shows the diversity of the vocabulary used by the source datasets. The right-hand side shows the average length of the texts. The large diversity our benchmark results from the merger of the four source datasets.

\begin{figure*}[!hbp]
    \centering
    \includegraphics[width=\textwidth]{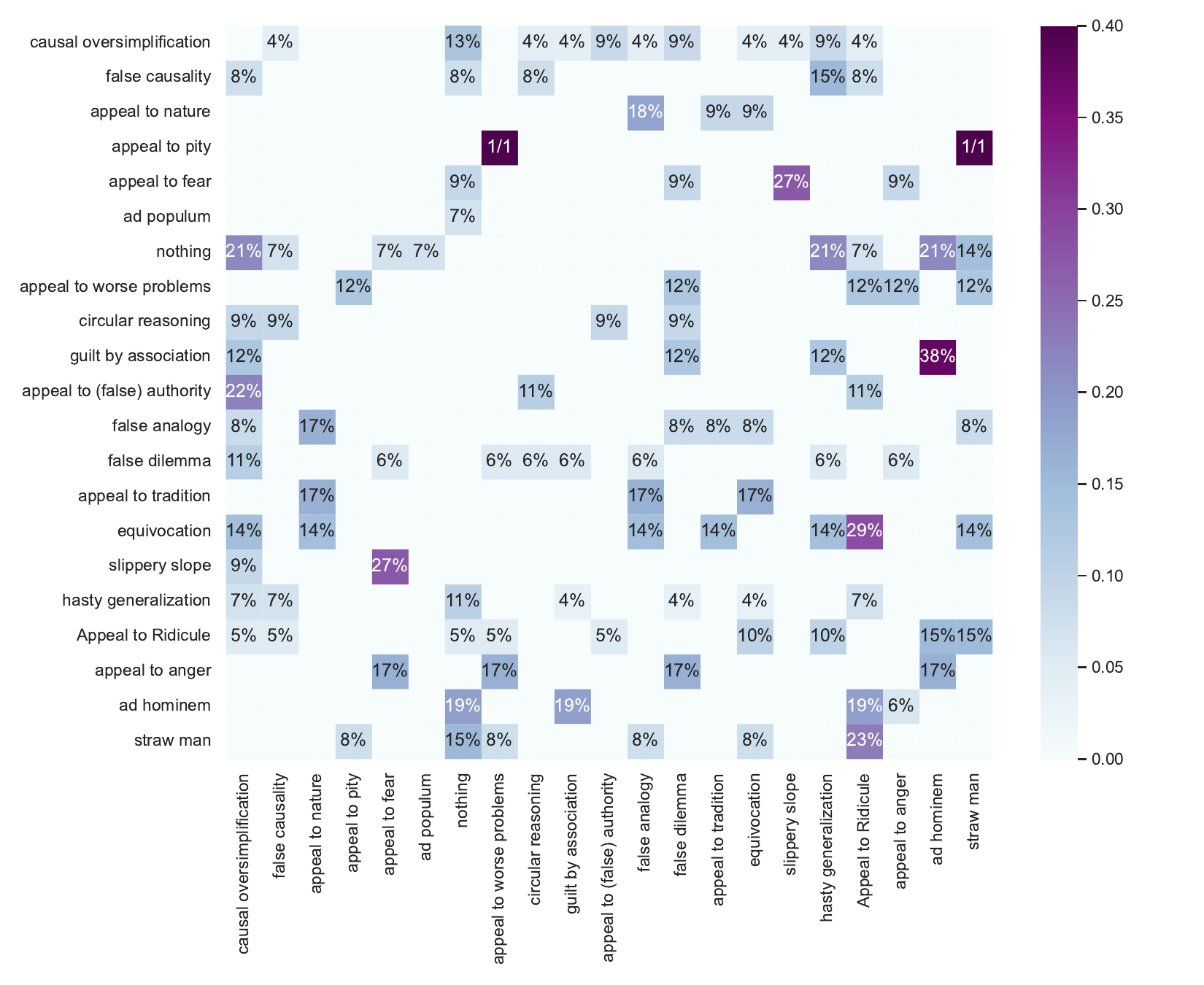}
    \caption{Co-occurrence of labels (frequency)}
    \label{fig:co-occurrence-freq}
\end{figure*}

Figure~\ref{fig:co-occurrence-freq} shows the co-occurrence frequency of each fallacy in the MAFALDA dataset. %

\begin{table}[ht]
\resizebox{\columnwidth}{!}{
    \begin{tabular}{lrr}
    \toprule
     & annotations & sources \\
    \midrule
    non-fallacious & 63 & 71 \\
    hasty generalization & 28 & 33 \\
    causal oversimplification & 23 & 0 \\
    Appeal to Ridicule & 20 &  0 \\
    false dilemma & 18 &  7 \\
    ad hominem & 16 &  8 \\
    nothing & 14 &  0 \\
    ad populum & 14 &  13 \\
    straw man & 13 &  0 \\
    false causality & 13 &  8 \\
    false analogy & 12 &  0 \\
    slippery slope & 11 &  6 \\
    appeal to fear & 11 &  0 \\
    appeal to nature & 11 &  10 \\
    circular reasoning & 11 &  10 \\
    appeal to (false) authority & 9 &  10 \\
    appeal to worse problems & 8 &  8 \\
    guilt by association & 8 &  0 \\
    equivocation & 7 & 1 \\
    appeal to tradition & 6 &  6 \\
    appeal to anger & 6 &  0 \\
    appeal to positive emotion & 3 &  0 \\
    tu quoque & 3 &  0 \\
    fallacy of division & 2 &  0 \\
    appeal to pity & 1 &  0 \\
    \midrule
    fallacy of relevance~\textsuperscript{*} & 0 &  2 \\
    intentional~\textsuperscript{*} & 0 &  1 \\
    appeal to emotion~\textsuperscript{*} & 0 &  10 \\
    \bottomrule
    \end{tabular}
    }
    \begin{minipage}{\textwidth}
        \footnotesize
        \textsuperscript{*} Fallacies not included in MAFALDA.\\
    \end{minipage}
    \caption{Number of spans for each fallacy: this table presents the distribution of fallacies in our dataset, comparing MAFALDA annotations with source annotations.}
    
    \label{tab:number_fallacy}
\end{table}

\section{Description of the Models}\label{app:models}

The computational budget was around 144 GPU hours for all models except GPT 3.5 and around \$2 for GPT 3.5 experiments. 
The GPU was an NVIDIA A 100. We used a temperature of 0.8.

\begin{description}
    \item GPT-3.5~\cite{brown2020language}, developed by OpenAI, is a transformer-based language model with 175 billion parameters, pre-trained on an extensive dataset encompassing a diverse range of texts. GPT-3.5 employs a technique known as Reinforcement Learning from Human Feedback (RLHF) for fine-tuning. 
    In this process, human trainers review and provide feedback on the model's outputs, ensuring the model responses are accurate and aligned with human judgment and values.

    \item Falcon~\cite{penedo2023refinedweb} is a large language model primarily pre-trained on the RefinedWeb -- a curated dataset extracted from CommonCrawl and refined for quality through filtering and deduplication. The model has two versions: 40B and 7B parameters.
    
    \item LLaMA-2~\cite{touvron2023llama}, developed by Meta, is a transformer-based language model pre-trained on 2 trillion tokens from various public sources. This model has multiple versions, including LLaMA 2-chat, tailored for dialogue applications. LLaMA-2 Instruct, another variant, has been fine-tuned using human instructions, LLaMA-2 generated instructions, and datasets like BookSum and Multi-document Question Answering. LLaMA-2 models come in different sizes, with parameters ranging from 7B to 70B.
    
    \item Vicuna~\cite{vicuna2023} is a model based on LLaMA, fine-tuned using a dataset comprising user conversations with ChatGPT. This model is available in two different sizes: 7B and 13B.
    
    \item Mistral~\cite{jiang2023mistral} is a 7B-parameter transformer-based model. It uses two attention mechanisms to improve inference speed and memory requirements: grouped-query attention (GQA) and sliding window attention (SWA). Specific details regarding the training data and hyperparameters are not disclosed.
    An alternative model version is also provided, fine-tuned to follow instructions. This refined model %
    was trained using publicly available instruction datasets from the Hugging Face repository.

    \item WizardLM~\cite{xu2023wizardlm} is a model based on LLaMa. It has been fine-tuned with a dataset comprising instructions that vary in complexity. The dataset was generated through a method known as Evol-Instruct, which systematically evolves simple instructions into more advanced ones. WizardLM is available in two sizes: 7B and 13B.
    
    \item Zephyr~\cite{tunstall2023zephyr} is a model based on Mistral and was fine-tuned on a variant of the UltraChat dataset, a synthetic dataset of dialogues generated by ChatGPT. Zephyr was further trained using the UltraFeedback dataset, which encompasses 64,000 ranked prompts and responses evaluated by GPT-4 to enhance its alignment.
    
\end{description}

\section{Level 2 Prompt} \label{app:prompt}
\textit{
Definitions:
\begin{itemize}
    \item An argument consists of an assertion called the conclusion and one or more assertions called premises, where the premises are intended to establish the truth of the conclusion. Premises or conclusions can be implicit in an argument.
    \item A fallacious argument is an argument where the premises do not entail the conclusion.
\end{itemize}
Text: "\{complete\_example\_input\}"\\
Based on the above text, determine whether the following sentence is part of a fallacious argument or not. If it is, indicate the type(s) of fallacy without providing explanations. The potential types of fallacy include:
\begin{itemize}
    \item appeal to positive emotion
    \item appeal to anger
    \item ...
    \item guilt by association
    \item tu quoque
\end{itemize}
Sentence: "\{sentence\_input\}"\\
Output:
}
\newline\newline
\textbf{An example and the generated output using GPT-3.5}:\\
\textit{
Definitions:
\begin{itemize}
    \item An argument consists of an assertion called the conclusion and one or more assertions called premises, where the premises are intended to establish the truth of the conclusion. Premises or conclusions can be implicit in an argument.
    \item A fallacious argument is an argument where the premises do not entail the conclusion.
\end{itemize}
Text: "I lost my phone in the living room, so it will always be in the living room when it is lost."\\
Based on the above text, determine whether the following sentence is part of a fallacious argument or not. If it is, indicate the type(s) of fallacy without providing explanations. The potential types of fallacy include:
\begin{itemize}
    \item appeal to positive emotion
    \item appeal to anger
    \item ...
    \item guilt by association
    \item tu quoque
\end{itemize}
Sentence: "I lost my phone in the living room, so it will always be in the living room when it is lost."\\
Output: This sentence is an example of the fallacy of hasty generalization.
}

\section{Metrics} \label{app:metrics}

\begin{figure*}[ht]
    \centering
    \includegraphics[width=\textwidth]{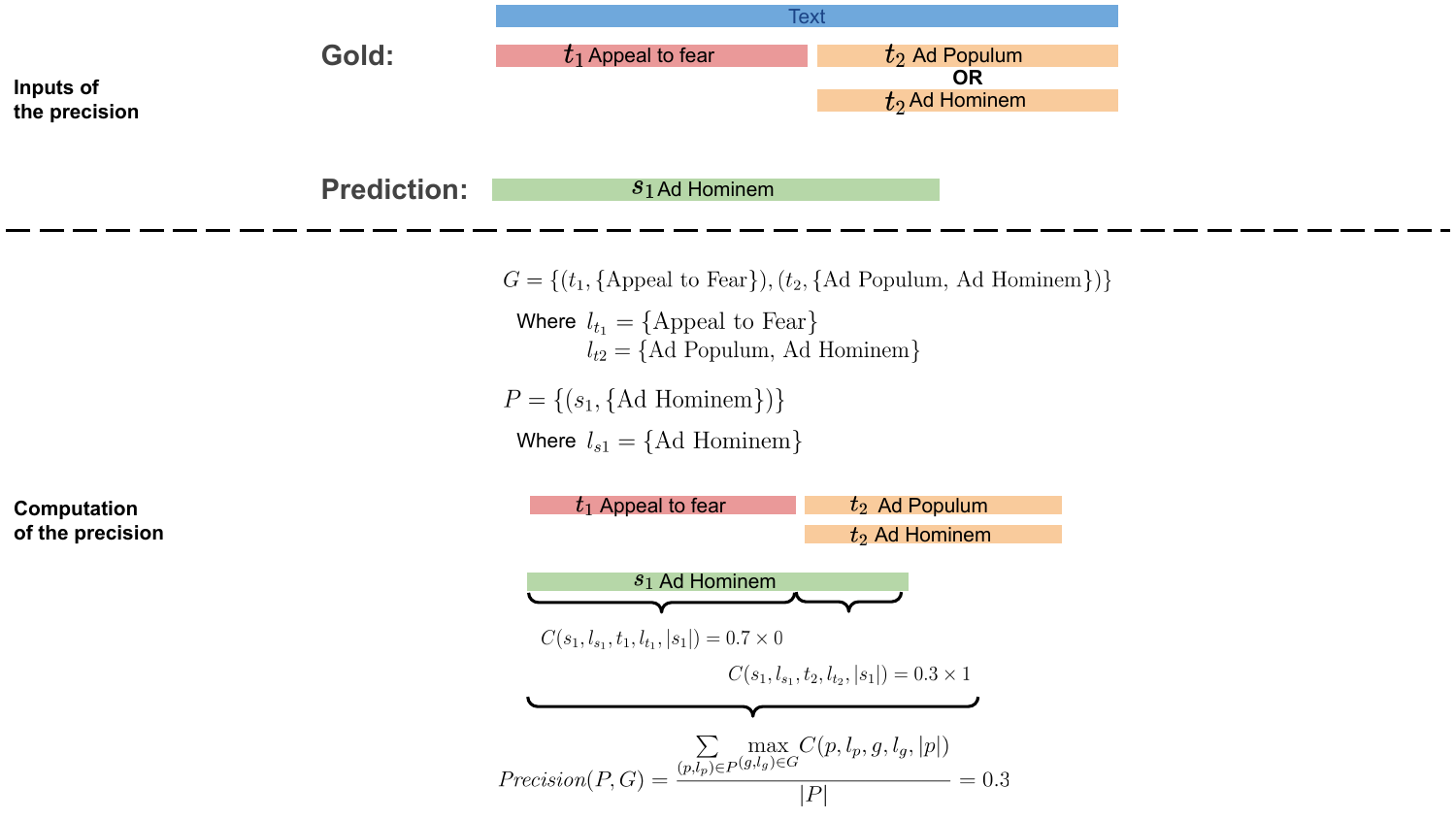}
    \caption{Example of Precision computation with alternatives.}
    \label{fig:precision_metric}
\end{figure*}

\begin{figure*}[ht]
    \centering
    \includegraphics[width=\textwidth]{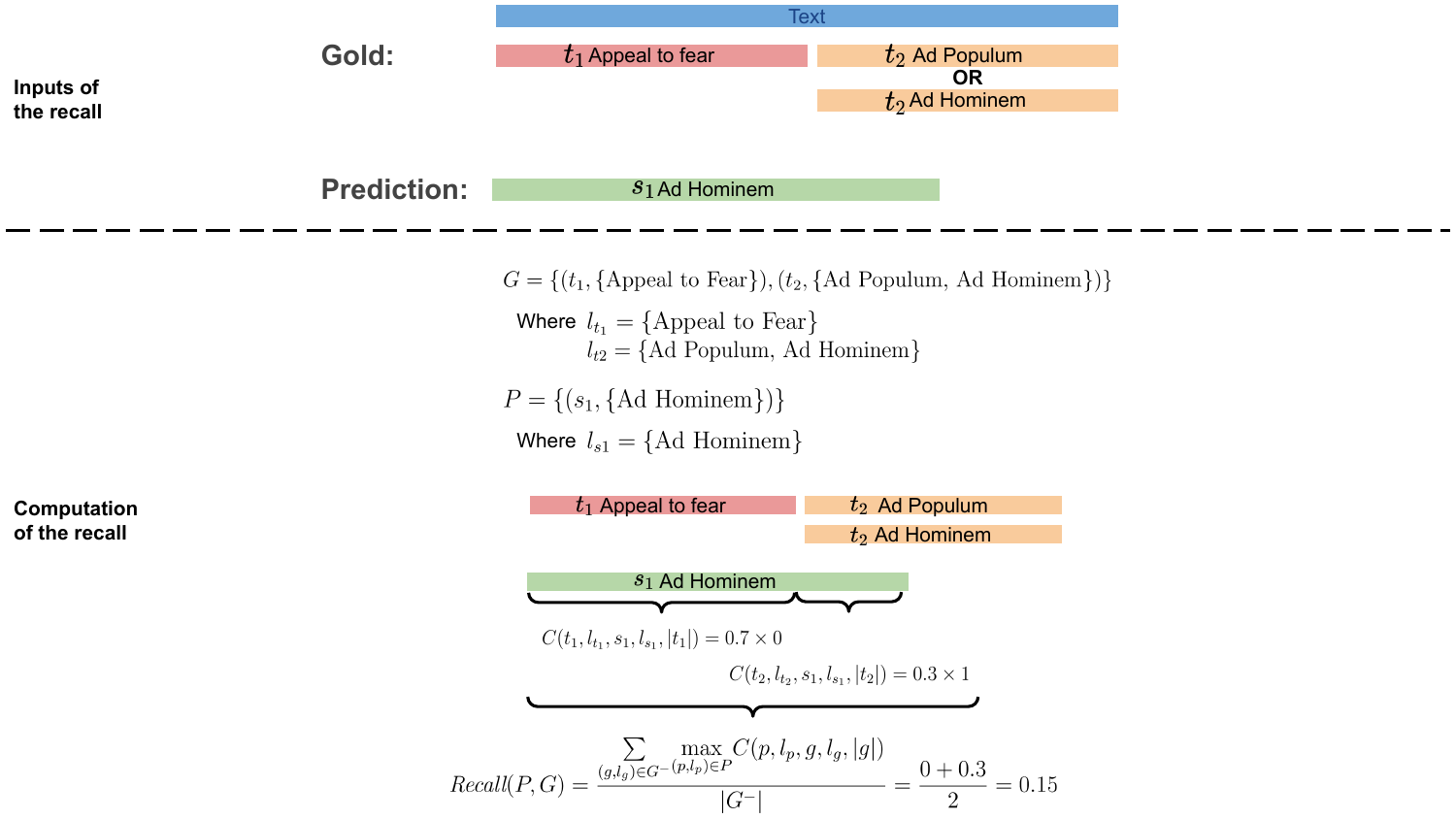}
    \caption{Example of Recall computation with alternatives.}
    \label{fig:recall_metric}
\end{figure*}

Figure~\ref{fig:precision_metric} and Figure~\ref{fig:recall_metric} show an example for the calculation of our precision and recall metrics are computed. We now prove some properties of our metrics. 

\begin{proposition}
Given a gold standard $G$, where each span comprises only a single sentence, and where each fallacy set contains only one element, and given a prediction $P$, where each span comprises only a single sentence,
our precision coincides with the standard precision.
\end{proposition}

\paragraph{Proof:} By definition, we have, for any spans $p$, $g$, and for any sets $l_p$, $l_g$:
\begin{align} &~~~ C(p, l_p, g, l_g, |p|)\\
= &~~~ \frac{|p \cap g|}{|p|} \times \delta(l_p, l_g)\\
= &~~~ \frac{|p \cap g|}{|p|} \times [l_p=l_g] 
\end{align}
If $p$ and $g$ are singleton spans, this boils down to
\begin{align}
= &~~~ [p=g] \times [l_p=l_g] \\
= &~~~ [p=g \wedge l_p=l_g]
\end{align}
Thus, we have, for any singleton span $s$ and any label $l$: 
\begin{align}
&~~~ [(s,\{l\}) \in G]\\
= &~~~ [\exists (s',\{l'\}) \in G: s'=s \wedge l'=l]\\
= &~~~ [\exists (s',\{l'\}) \in G: C(s, l, s', l', |s|)=1]\\ 
= &~~~ \max_{(s',\{l'\}) \in G} C(s, l, s', l', |s|)
\end{align}
This entails that the number of true positives (TP) is
\begin{align}
&~~~ |\{(s,l) \in P \mid (s, \{l\}) \in G\}|\\
= &~~~ \sum_{(s,l)\in P} [(s, \{l\}) \in G]\\
= &~~~ \sum_{(s,l)\in P} \max_{(s',\{l'\}) \in G} C(s, l, s', l', |s|) \label{eqv:tp}
\end{align}
The standard precision is the ratio of true positives (TP) out of the sum of true positives and false positives (FP):
\begin{align} & \text{Standard Precision} \\
= &~~~ \frac{\text{TP}}{\text{TP} + \text{FP}} \end{align}
With $|P|=\text{TP}+\text{FP}$ and Equation~\ref{eqv:tp}, this is equivalent to
\begin{align} = &~~~ \frac{\sum_{(s,l)\in P} \max_{(s',\{l'\}) \in G} C(s, l, s', l', |s|)}{|P|}\nonumber \end{align}
\qed

\newcommand{\ignore}[1]{}
\ignore{

 \[\textit{Precision}(P, G) = \frac{\sum\limits_{(p, l_p)\in P}\!
\max\limits_{(g, l_g)\in G}\!C(p, l_p, g, l_g, |p|)}{|P|}\]

\textbf{Given Condition}: Each span annotation is single-sentenced with one label.

\noindent Under this condition, for each annotation \( (s,l) \in P \), the score is either 1 (True Positive) if \( (s,l) \in G \), or 0 (False Positive) if it does not. Therefore:
\[ \sum\limits_{(p, l_p)\in P}\!
\max\limits_{(g, l_g)\in G}\!C(p, l_p, g, l_g, |p|) = \text{True Positives} \]

\begin{align*}
    |P| &= \text{Total Predictions}\\
    &= \text{True Positives} + \text{False Positives}
\end{align*}

\begin{align*} 
Precision(P, \mathcal{G}) &= \frac{\text{TP}}{\text{TP} + \text{FP}}\\
&= \text{Standard Precision}
\end{align*}

}

\begin{proposition}
Given a gold standard $G$, where each span comprises only a single sentence, and where each fallacy set contains only one element, and given a prediction $P$, where each span comprises only a single sentence, our recall coincides with the standard recall.
\end{proposition}
\paragraph{Proof:} 
As previously, for any $p$, $g$ that are singleton spans, we have:
\begin{align} &~~~ C(p, l_p, g, l_g, |g|)\\
= &~~~ [p=g \wedge l_p=l_g]
\end{align}
Thus, we have, for any singleton span $s$ and any label $l$: 
\begin{align}
&~~~ [(s',l') \in P]\\
= &~~~ [\exists (s, l) \in P: s=s' \wedge l=l']\\
= &~~~ [\exists (s, l) \in P: C(s, l, s', l', |s'|)=1]\\ 
= &~~~ \max_{(s, l) \in P} C(s, l, s', l', |s'|)
\end{align}
This entails that the number of true positives (TP) is
\begin{align}
&~~~ |\{(s', \{l'\}) \in G^- \mid (s', l') \in P\}|\\
= &~~~ \sum_{(s',\{l'\})\in G^-} [(s', l') \in P]\\
= &~~~ \sum_{(s',\{l'\})\in G^-} \max_{(s, l) \in P} C(s, l, s', l', |s'|) \label{eqv:tp2}
\end{align}
The standard recall is the ratio of true positives (TP) out of the sum of true positives and false negatives (FN):
\begin{align} & \text{Standard Recall} \\
= &~~~ \frac{\text{TP}}{\text{TP} + \text{FN}} \end{align}
With $|G^-|=\text{TP}+\text{FN}$ and Equation~\ref{eqv:tp2}, this is equivalent to
\begin{align} = &~~~ \frac{\sum_{(s,\{l\})\in G^-} \max_{(s', l') \in P} C(s, l, s', l', |s|)}{|G^-|}\nonumber \end{align}
\qed

\begin{proposition}
In cases where a system predicts multiple labels for a single span, and the corresponding gold standard also contains multiple alternative labels for that span, the system's recall does not increase with the number of correctly predicted labels. Our recall formula ensures that multiple predictions for the same span do not artificially inflate the recall metric.
\end{proposition}

\noindent\textbf{Proof:} By our definition, we have, for any prediction $P$ and any gold standard $G$:
\begin{align}\textit{Recall}(P, G)\!=\!\!\frac{\sum\limits_{(g, l_g)\in G^-}\!
\max\limits_{(p, l_p)\in P}\!C(p, l_p, g, l_g, |g|)}{|G^-|}\end{align}

\noindent By definition, we have, for any spans $p$, $g$, and for any sets $l_p$, $l_g$:
\begin{align} &~~~ C(p, l_p, g, l_g, |g|)\\
= &~~~ \frac{|p \cap g|}{|g|} \times \delta(l_p, l_g)
\end{align}
If $p$ and $g$ are the same spans, this boils down to
\begin{align}
= &~~~ \delta(l_p, l_g)\\
= &~~~ [l_p \in l_g \neq \emptyset]
\end{align}

\noindent The $\max$ operation ensures that the score contribution for the span in the recall is based on the best match between predicted labels and gold standard labels, capped at 1 regardless of the number of labels correctly predicted. So even if multiple labels in $l_p$ for $p$ in $P$ match with different alternative labels in $l_g$ in $G$, the contribution to the recall for the span remains 1.

\begin{align}
\max(C(p1, l_{p1}, g, l_{g}, |g|),\\ C(p2, l_{p2}, g, l_{g}, |g|),\\ ,...,\\ C(pn, l_{pn}, g, l_{g}, |g|))\\
&= 1
\end{align}
\qed

\subsection{Metric equivalence} \label{sec:metric_equiv}

The metric proposed in this paper is similar to the metric proposed in \cite{martinoFineGrainedAnalysisPropaganda2019}. This metric supposes that there is no overlap of spans with the same label. However, such spans are very frequent in a multi-level taxonomy, when evaluating Levels 0 and 1. Consider the following example:

\begin{myexamplebox}
\textit{You are a liar. Therefore you are wrong.}
\end{myexamplebox}

\noindent In this example, there is only one \adhominem, which is a \textit{Fallacy of Credibility} on Level 1. Now assume that the model outputs:
(You are a liar, \adhominem), (You are a liar, therefore you are wrong, \tuquoque). Using the recall from \cite{martinoFineGrainedAnalysisPropaganda2019}, and $G=\{([0,40], \textsc{Credibility)\}}$, $P=\{([0,10], \textsc{Credibility}), ([0,40], \textsc{Credibility})\}$ we get the following recall:
\begin{align*}
Recall_m(P, G)&= \frac{1}{|G|}\sum_{p\in P, g\in G}{ C_m(p,g,|g|)} \\
            &= \frac{1}{1} * (\frac{10}{40}*1+\frac{40}{40}*1) \\
            &= 1.25
\end{align*}
\noindent Instead of computing one score for each element of $G$ as it would be expected for the recall, the metrics is computing all scores between all spans with the same label. We thus get a score larger than one.

Hence, we propose to sum only the best match for each element of $G$. 

\begin{align*}
\\
\textit{Recall}(P, G)\!&=\!\!\frac{\sum\limits_{(g, l_g)\in G^-}\!\max\limits_{(p, l_p)\in P}\!C(p, l_p, g, l_g, |g|)}{|G^-|}\\
            &= \frac{1}{1} * (max(\frac{10}{40}, \frac{40}{40})) \\
            &= 1
\end{align*}

\begin{proposition}
Given a gold standard $G$, where for each span there are no alternatives, and there is only one span from a prediction $P$ that overlaps with one span from the gold standard, our recall metric $Recall(P, G)$ coincides with \citet{martinoFineGrainedAnalysisPropaganda2019}'s $Recall_m(P, G)$.
\end{proposition}

\noindent \textbf{Proof:} 
 Given a gold standard $G$ with no alternatives so:
\begin{align}
    G=G^-
\end{align}

\noindent By definition, we have, for any spans $p$, $g$, and for any sets $l_p$, $l_g$:
\begin{align} &~~~ C(p, l_p, g, l_g, |g|)\\
= &~~~ \frac{|p \cap g|}{|g|} \times \delta(l_p, l_g)\\
= &~~~ \frac{|p \cap g|}{|g|} \times [l_p=l_g] 
\end{align}

\noindent In case of \citet{martinoFineGrainedAnalysisPropaganda2019}'s comparison score $C_m$, we have:
\begin{align}
    &~~~C_m(p, g, |g|)\\
    = &~~~\frac{|p \cap g|}{|g|} \times \delta(l(p), l(g))
\end{align}

\noindent In \citet{martinoFineGrainedAnalysisPropaganda2019}'s comparison score $C_m$, $l$ represents a labeling function, so in this case:
\begin{align}
    &~~~C_m(p, g, |g|)\\
    = &~~~\frac{|p \cap g|}{|g|} \times \delta(l(p), l(g))\\
    = &~~~ \frac{|p \cap g|}{|g|} \times [l_p=l_g]\\
    = &~~~ C(p, l_p, g, l_g, |g|)
\end{align}

\noindent For a given label, there is either no prediction span that overlaps or a unique prediction span $p$ that overlaps with a gold standard span $g$, such that their overlap and label agreement is non-zero:
If there is no span $p$ in $P$ that overlaps, then both recalls are equal to zero. 
The other case is:
\begin{align}
\forall (g, l) &\in G,  \exists ! (p, l) \in P:  \frac{|p \cap g|}{|g|} \times [l=l]\\
= &~~~ C(p, l, g, l, |g|)\\
= &~~~ C_m(p, g, |g|)\\
> &~~~ 0
\end{align}

\noindent This implies that for each gold standard annotation, the maximum score of $C$ between each annotation in the gold standard and the annotations in the prediction is achieved exactly once:

\begin{align}
\forall (g, l_g) \in G, &\max\limits_{(p, l_p)\in P}\!C(p,l_p,g,l_g,|g|)\\ &~~~= C_m(p, g, |g|)
\end{align}

\noindent Additionally, for each gold standard annotation, the sum of  scores $C$ across all predictions equals the maximum score  $C$, since only one prediction per annotation contributes a non-zero score, so:

\begin{align}
\forall (g, l_g) \in G, &\sum\limits_{(p, l_p)\in P} C(p,l_p,g,l_g,|g|) \\ &~~~= C_m(p, g, |g|)
\end{align}

\noindent Finally,

\begin{align} 
& \frac{\sum\limits_{(p, l_p) \in P, (g, l_g) \in G}{ C(p,l_p,g,l_g,|g|)}}{|G|}\\ &= \frac{1}{|G|}\sum\limits_{p\in P, g\in G} C_m(p, g, |g|)
\end{align}

\begin{align}
& \frac{\sum\limits_{(g, l_g)\in G^-}\!
\max\limits_{(p, l_p)\in P}\!C(p,l_p,g,l_g,|g|)}{|G^-|}\\ &= \frac{1}{|G|}\sum\limits_{p\in P, g\in G}  C_m(p, g, |g|) \\
\end{align}

We can conclude that: 
\begin{align}
    Recall_m(P,G)=Recall(P,G) 
\end{align}
\qed

A similar demonstration can be done for precision.

\begin{figure}[!ht]
    \centering
    \includegraphics[width=\columnwidth]{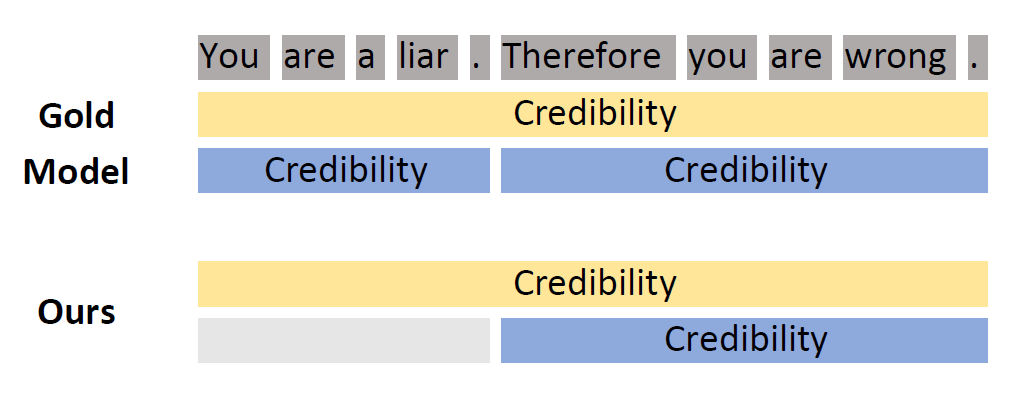}
    \caption{Illustration of the difference between our metric and the one from \citep{martinoFineGrainedAnalysisPropaganda2019}. %
    In Martino's metric, both annotated spans count, and we get a recall of 1. Our metric counts only the largest overlapping span (light blue), and gives a recall of 0.6.}
    \label{fig:diff_metrics}
\end{figure}

There is another difference between our metrics and the original one: If two disjoint spans from the prediction overlap with the gold standard, in \cite{martinoFineGrainedAnalysisPropaganda2019}'s metrics, they both contribute to the score. In our metrics, only the best match contributes to the score. Our metric thus rewards models that output the correct span without splitting it into multiple spans (see Figure~\ref{fig:diff_metrics}).

\section{Results} \label{app:results}

Table~\ref{tab:result} displays the F1-scores for our experiments on both the complete dataset and the subset from the user study, across all three task levels, with the best scores for each level highlighted in bold. For more in-depth analysis, Table~\ref{tab:detailed_results} provides detailed results, including Recall, Precision, and F1-score for Levels 0, 1, and 2 (referenced as Tables~\ref{tab:result_lvl0}, \ref{tab:result_lvl1}, and \ref{tab:result_lvl2}, respectively), along with corresponding data from the user study (Table~\ref{tab:user_study_details}).

\begin{table*}[!ht]
    \centering
    \resizebox{\textwidth}{!}{

     \begin{tabular}{lcccccc}
        \toprule
        & \multicolumn{3}{c}{MAFALDA} & \multicolumn{3}{c}{Sample of MAFALDA} \\
        \cmidrule(lr){2-4} \cmidrule(lr){5-7}
        Model & F1 Level 0 & F1 Level 1 & F1 Level 2 & F1 Level 0 & F1 Level 1 & F1 Level 2 \\
        \midrule
        Baseline random & 0.435 & 0.061 & 0.010 & 0.211 & 0.013& 0.004 \\
        \midrule
        Falcon 7B &  0.397 & 0.130 & 0.022 & 0.274 & 0.099 & 0.019\\
        LLaMA2 Chat 7B & 0.572 & 0.114 & 0.068  & 0.356 & 0.065 & 	0.030\\
        LLaMA2 Chat 13B & 0.549 & 0.160 & 0.096& 0.364 & 0.103 & 0.043\\
        LLaMA2  7B&  0.492 & 0.148 & 0.038 & 	0.347 & 0.145 & 	0.037\\
        LLaMA2 13B & 0.458 & 0.129 & 0.039& 0.309 & 0.109 & 0.003\\
        Mistral Instruct 7B &0.536 & 0.144 & 0.069  & 0.404 & 0.089 & 0.004\\
        Mistral  7B & 0.450 & 0.127 & 0.044 & 0.393 & 0.102 & 0.017\\
        Vicuna 7B &  0.494 & 0.134 & 0.051 & 0.258 & 0.061 & 0.049\\
        Vicuna 13B &  0.557 & 0.173 & 0.100 & 0.293 & 0.121 & 0.032\\
        WizardLM 7B & 0.490 & 0.087 & 0.036 & 0.233 & 0.036 & 0.0\\
        WizardLM 13B & 0.520 & 0.177 & 0.093 & 0.246 & 0.123 &0.021\\
        Zephyr 7B &0.524 & 0.192 & 0.098& 0.312 & 0.109 & 0.025\\
        \midrule
        GPT 3.5 175B & \textbf{0.627} & \textbf{0.201} & \textbf{0.138} & 0.338 & 0.095& 0.034\\
        \midrule
        Avg. Human & - & - & - & \textbf{0.749} & \textbf{0.352} & \textbf{0.186}\\
        \bottomrule
    \end{tabular}
    }

    \caption{Performance results of different models across different granularity levels in a zero-shot setting. The right part concerns only the user study with a subsample of 20 texts from MAFALDA. The best results for each level are highlighted in bold.\label{tab:result}}
\end{table*}

\begin{table*}[ht]
\centering
\begin{subtable}[]{\columnwidth}
    \centering
    \resizebox{\columnwidth}{!}{
    \begin{tabular}{lrrr}
        \toprule
        Model & Precision Level 0 & Recall Level 0 & F1 Level 0 \\
        \midrule
        Falcon 7B & 0.427 & 0.655 & 0.397 \\
        LLAMA2 Chat 7B& 0.506 & 0.837 & 0.572 \\
        LLAMA2 7B& 0.456 & 0.758 & 0.492 \\
        Mistral Instruct 7B & 0.570 & 0.651 & 0.536 \\
        Mistral  7B & 0.444 & 0.691 & 0.450 \\
        Vicuna 7B & 0.529 & 0.628 & 0.494 \\
        WizardLM 7B & 0.565 & 0.567 & 0.490 \\
        Zephyr 7B & 0.489 & 0.765 & 0.524 \\
        LLaMA2 Chat 13B & 0.493 & 0.793 & 0.549 \\
        LLaMA2 13B & 0.433 & 0.739 & 0.458 \\
        Vicuna 13B & 0.591 & 0.670 & 0.557 \\
        WizardLM 13B & 0.523 & 0.756 & 0.520 \\
        GPT 3.5 175B & 0.701 & 0.669 & 0.627 \\
            \bottomrule
    \end{tabular}
        }
    \caption{Performance results for Level 0\label{tab:result_lvl0} on MAFALDA}
\end{subtable}%
\begin{subtable}[]{\columnwidth}
    \centering
    \resizebox{\columnwidth}{!}{
    \begin{tabular}{lrrr}
    \toprule
    Model & Precision Level 1 & Recall Level 1 & F1 Level 1 \\
    \midrule
    Falcon 7B& 0.134 & 0.164 & 0.130 \\
    LLAMA2 Chat 7B& 0.134 & 0.136 & 0.114 \\
    LLAMA2 7B& 0.158 & 0.185 & 0.148 \\
    Mistral Instruct 7B & 0.176 & 0.152 & 0.144 \\
    Mistral  7B & 0.136 & 0.159 & 0.127 \\
    Vicuna 7B & 0.161 & 0.146 & 0.134 \\
    WizardLM 7B & 0.121 & 0.093 & 0.087 \\
    Zephyr 7B & 0.207 & 0.230 & 0.192 \\
    LLaMA2 Chat 13B & 0.173 & 0.183 & 0.160 \\
    LLaMA2 13B & 0.140 & 0.151 & 0.129 \\
    Vicuna 13B & 0.200 & 0.191 & 0.173 \\
    WizardLM 13B & 0.193 & 0.205 & 0.177 \\
    GPT 3.5 175B & 0.233 & 0.203 & 0.201 \\
    \bottomrule
    \end{tabular}
    }
 \caption{Performance results for Level 1\label{tab:result_lvl1} on MAFALDA}
\end{subtable}
\bigskip
\vspace{3mm}
\begin{subtable}[]{\columnwidth}
    \centering
    \resizebox{\columnwidth}{!}{
    \begin{tabular}{lrrr}
    \toprule
    Model & Precision Level 2 & Recall Level 2 & F1 Level 2 \\
    \midrule
    Falcon 7B& 0.016 & 0.078 & 0.022 \\
    LLAMA2 Chat 7B& 0.070 & 0.095 & 0.068 \\
    LLAMA2 7B& 0.038 & 0.073 & 0.038 \\
    Mistral Instruct 7B & 0.086 & 0.076 & 0.069 \\
    Mistral  7B & 0.046 & 0.072 & 0.044 \\
    Vicuna 7B & 0.062 & 0.067 & 0.051 \\
    WizardLM 7B & 0.056 & 0.041 & 0.036 \\
    Zephyr 7B & 0.090 & 0.145 & 0.098 \\
    LLaMA2 Chat 13B & 0.101 & 0.122 & 0.096 \\
    LLaMA2 13B & 0.037 & 0.068 & 0.039 \\
    Vicuna 13B & 0.115 & 0.118 & 0.100 \\
    WizardLM 13B & 0.088 & 0.134 & 0.093 \\
    GPT 3.5 175B & 0.162 & 0.138 & 0.138 \\

    \bottomrule
    \end{tabular}
    }
 \caption{Performance results for Level 2\label{tab:result_lvl2} on MAFALDA}
\end{subtable}
\begin{subtable}[]{\columnwidth}
    \centering
    \resizebox{0.65\columnwidth}{!}{
    \begin{tabular}{llrrr}
    \toprule
    & Model & Precision & Recall & F1 \\
    \midrule
    \multirow{5}{*}{Level 0} & user1 & 0.732 & 0.847 & 0.760 \\
    & user2 & 0.785 & 0.892 & 0.821 \\
    & user4 & 0.728 & 0.809 & 0.728 \\
    & user5 & 0.704 & 0.767 & 0.694 \\
    & Average & 0.737 & 0.829 & 0.749 \\
    \midrule
    \multirow{5}{*}{Level 1} & user1 & 0.326 & 0.342 & 0.322 \\
    & user2 & 0.399 & 0.402 & 0.397 \\
    & user4 & 0.311 & 0.364 & 0.319 \\
    & user5 & 0.375 & 0.394 & 0.371 \\
    & Average & 0.353 & 0.376 & 0.352 \\
    \midrule
    \multirow{5}{*}{Level 2} & user1 & 0.192 & 0.248 & 0.204 \\
    & user2 & 0.162 & 0.172 & 0.164 \\
    & user4 & 0.186 & 0.239 & 0.194 \\
    & user5 & 0.170 & 0.211 & 0.180 \\
    & Average & 0.177 & 0.217 & 0.186 \\
    \bottomrule
    \end{tabular}
    }
    \caption{Performances results for the user Study}
    \label{tab:user_study_details}
\end{subtable}
\caption{Detailed results of the experiments, including Recall, Precision, and F1-score, for each level, for models and each user of the user study.%
\label{tab:detailed_results}} 
\end{table*}

\section{Error Analysis} \label{app:error_analysis}
\begin{figure*}[ht]
    \centering
    \begin{subfigure}[b]{0.45\textwidth}
        \includegraphics[width=\textwidth]{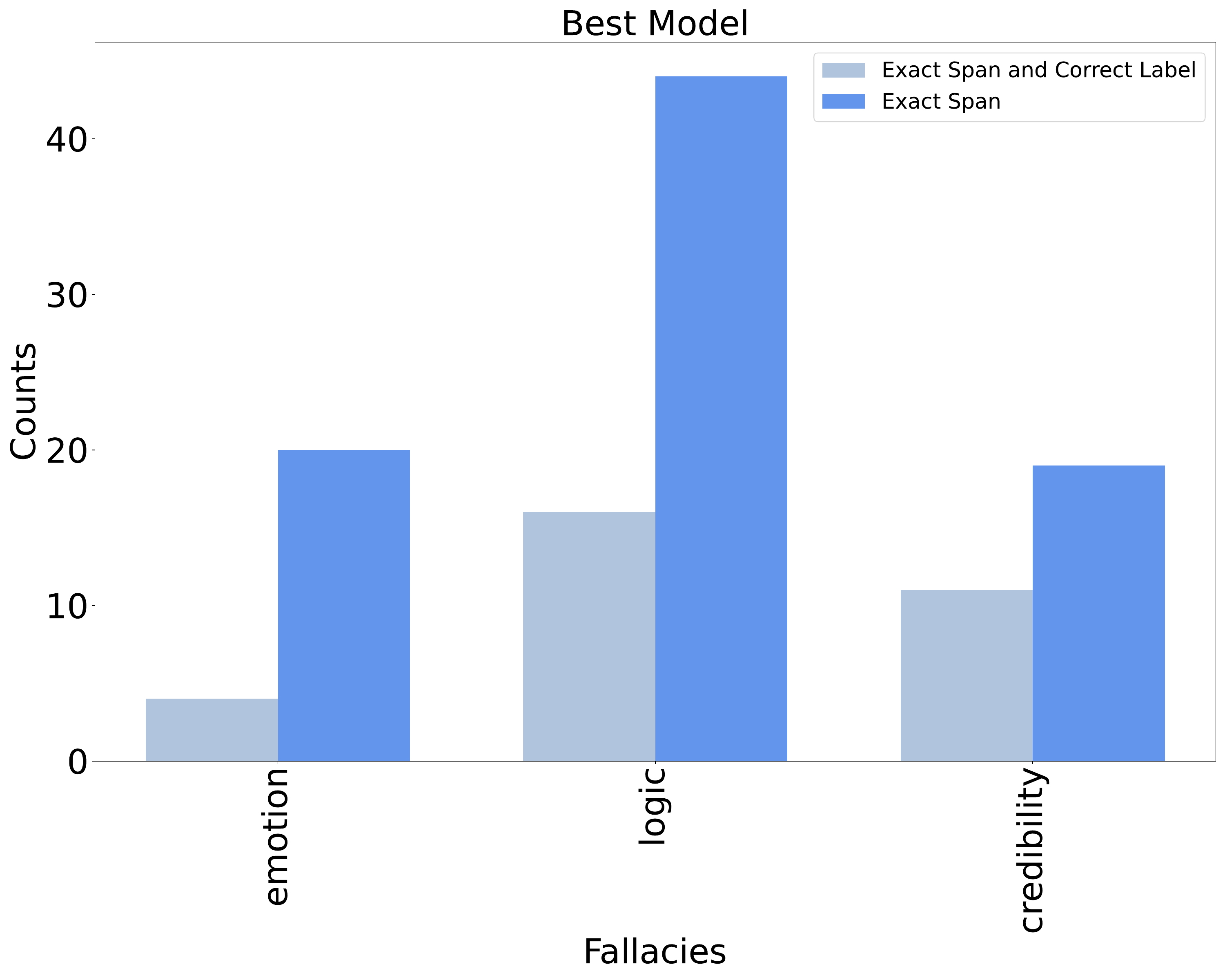}
        \caption{Accuracy of fallacy labeling for spans that exactly match the gold standard at Level 1 for the \textbf{best model}}
        \label{fig:correct_fallacy_best_model_level_1}
    \end{subfigure}
    \hfill
    \begin{subfigure}[b]{0.45\textwidth}
        \includegraphics[width=\textwidth]{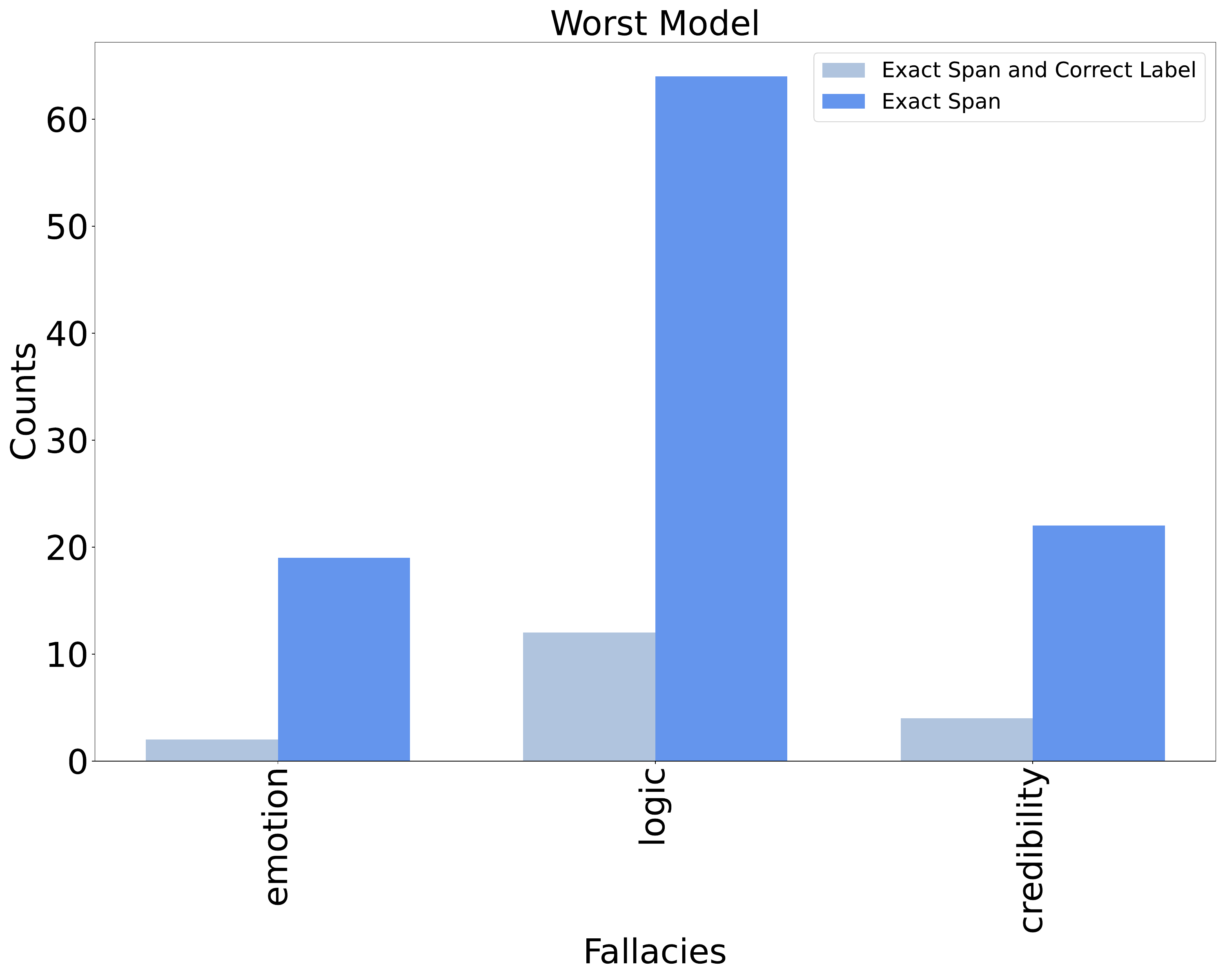}
        \caption{Accuracy of fallacy labeling for spans that exactly match the gold standard at Level 1 for the \textbf{worst model}}
        \label{fig:correct_fallacy_worst_model_level_1}
    \end{subfigure}

    \caption{
    Accuracy of fallacy labeling for spans that \textbf{exactly match} the gold standard at Level 1 for the best and worst models. \emph{Exact Span} corresponds to the number of spans correctly identified by the model, \emph{Exact Span and Correct Label} corresponds to the number of correctly labeled spans out of the correctly identified spans.
    }
    \label{fig:correct_fallacy_models_level_1}
\end{figure*}

\begin{figure*}[ht]
    \centering
    \begin{subfigure}[b]{0.45\textwidth}
        \includegraphics[width=\textwidth]{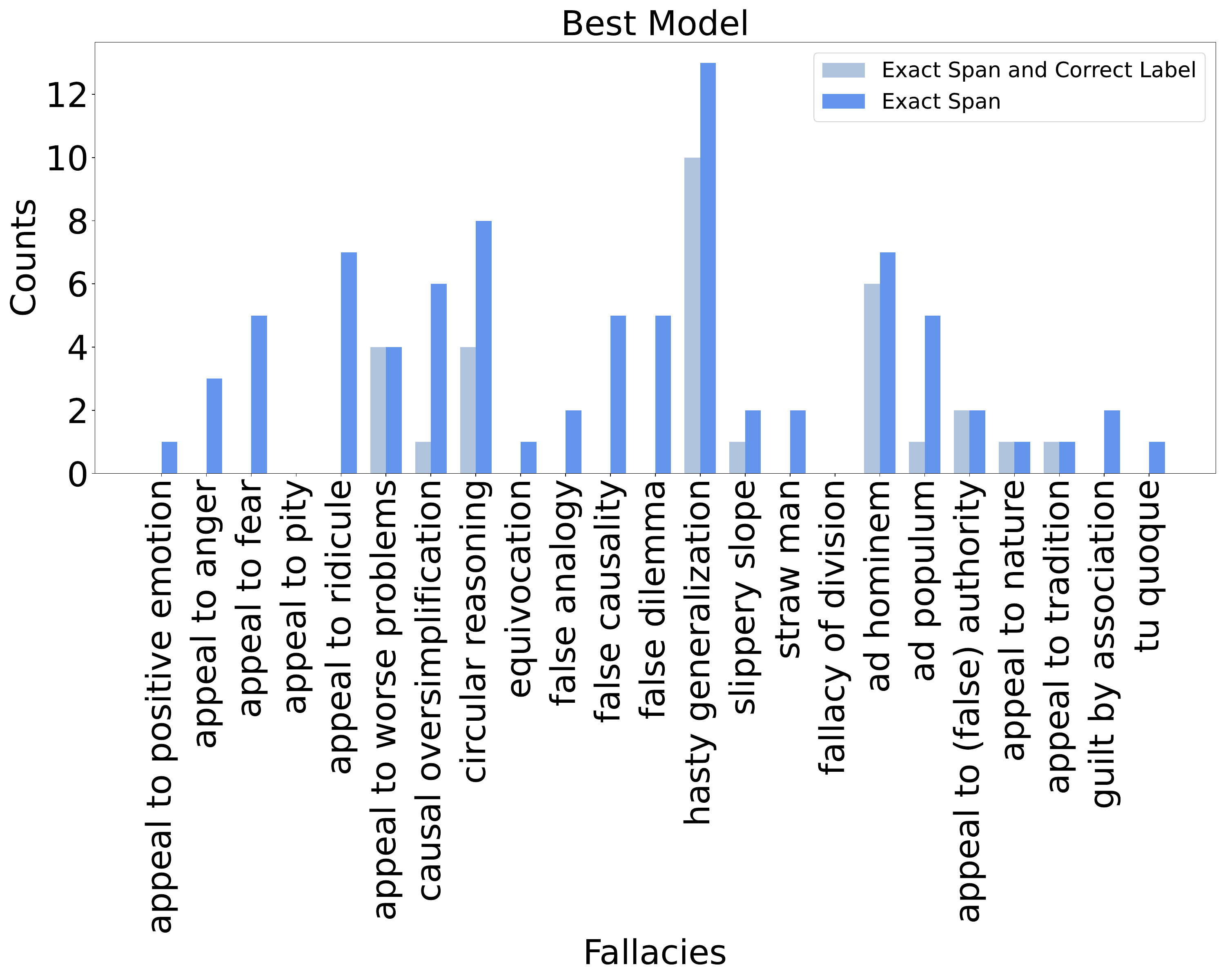}
        \caption{Accuracy of fallacy labeling for spans that exactly match the gold standard at Level 2 for the \textbf{best model}}
        \label{fig:correct_fallacy_best_model_level_2}
    \end{subfigure}
    \hfill
    \begin{subfigure}[b]{0.45\textwidth}
        \includegraphics[width=\textwidth]{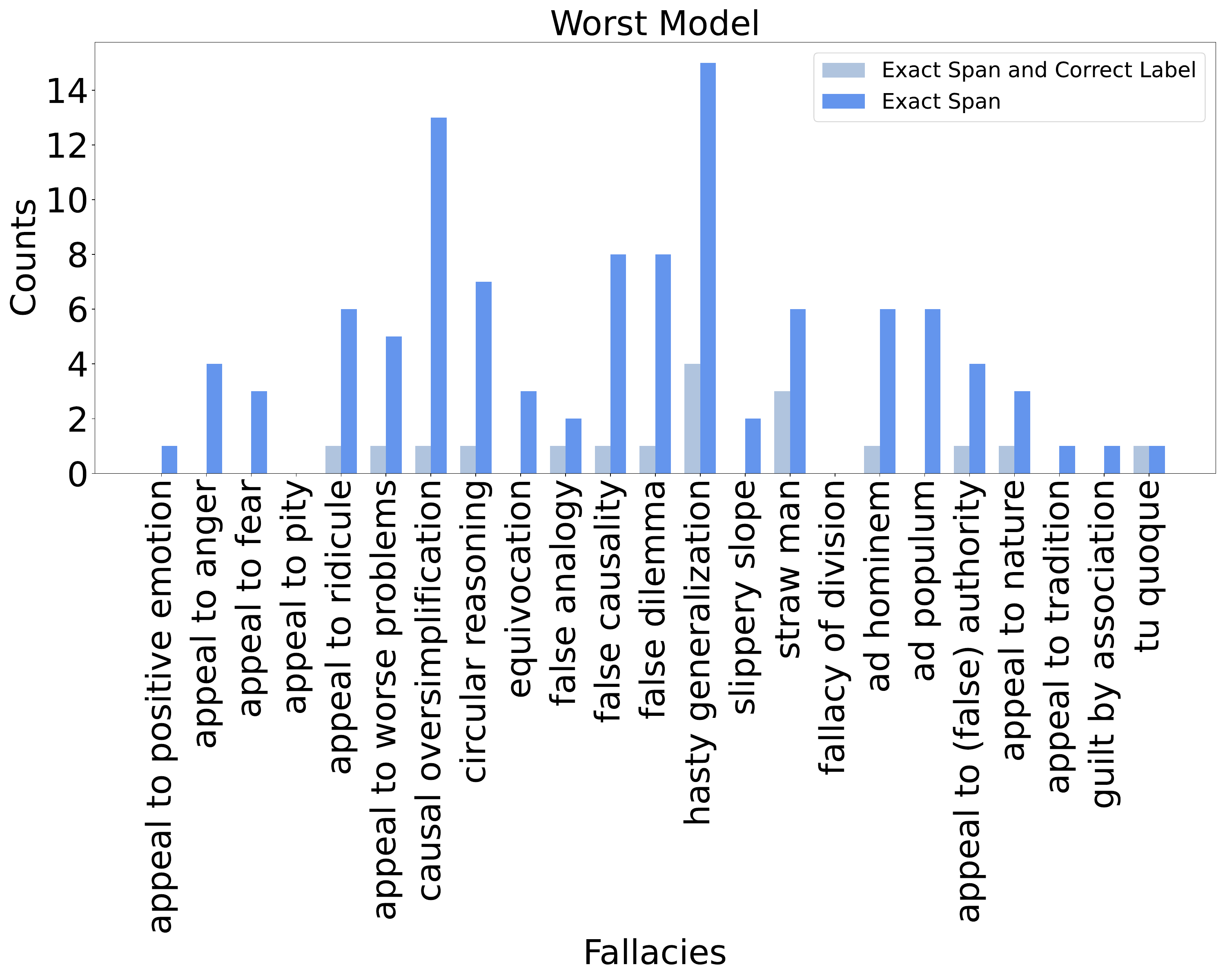}
        \caption{Accuracy of fallacy labeling for spans that exactly match the gold standard at Level 2 for the \textbf{worst model}}
        \label{fig:correct_fallacy_worst_model_level_2}
    \end{subfigure}

    \caption{Accuracy of fallacy labeling for spans that \textbf{exactly match} the gold standard at Level 2 for the best and worst models. \emph{Exact Span} corresponds to the number of spans correctly identified by the model, \emph{Exact Span and Correct Label} corresponds to the number of correctly labeled spans out of the correctly identified spans.}
    \label{fig:correct_fallacy_models_level_2}
\end{figure*}

\begin{figure*}[ht]
    \centering
    \begin{subfigure}[b]{0.45\textwidth}
        \includegraphics[width=\textwidth]{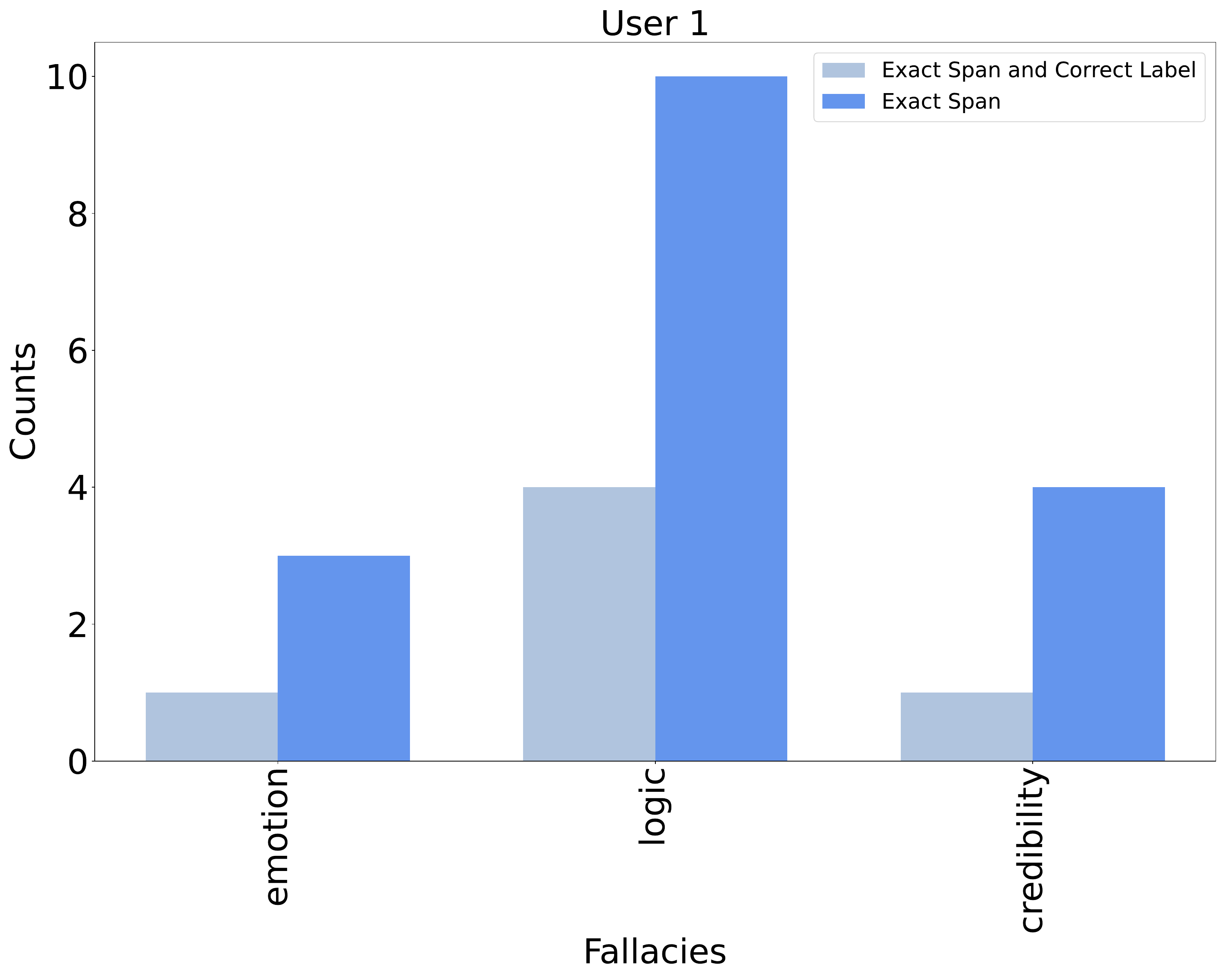}
        \caption{Accuracy of fallacy labeling for spans that exactly match the gold standard at Level 1 for the \textbf{User 1}}
        \label{fig:correct_fallacy_user_1_level_1}
    \end{subfigure}
    \hfill
    \begin{subfigure}[b]{0.45\textwidth}
        \includegraphics[width=\textwidth]{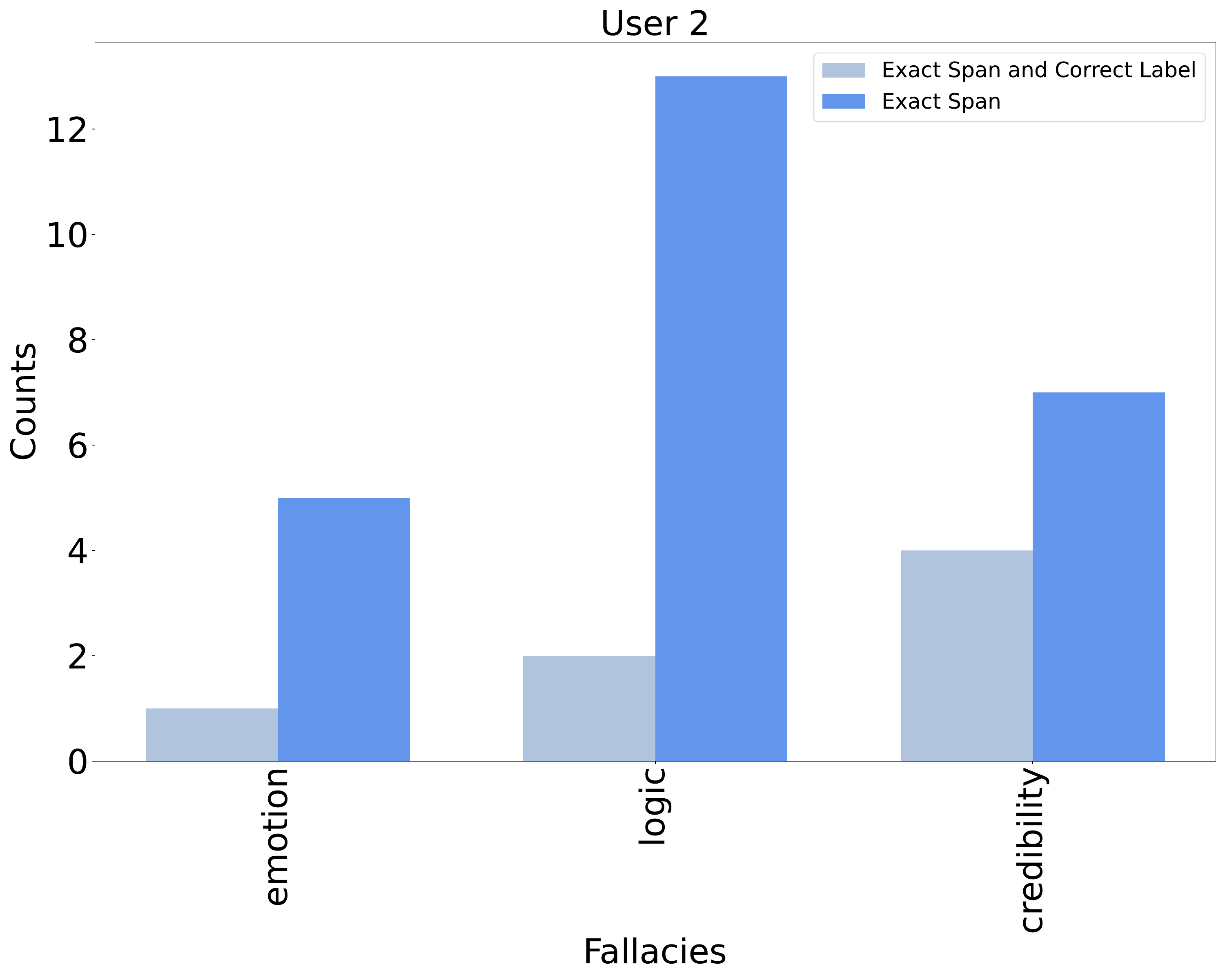}
        \caption{Accuracy of fallacy labeling for spans that exactly match the gold standard at Level 1 for the \textbf{User 2}}
        \label{fig:correct_fallacy_user_2_level_1}
    \end{subfigure}
    \begin{subfigure}[b]{0.45\textwidth}
        \includegraphics[width=\textwidth]{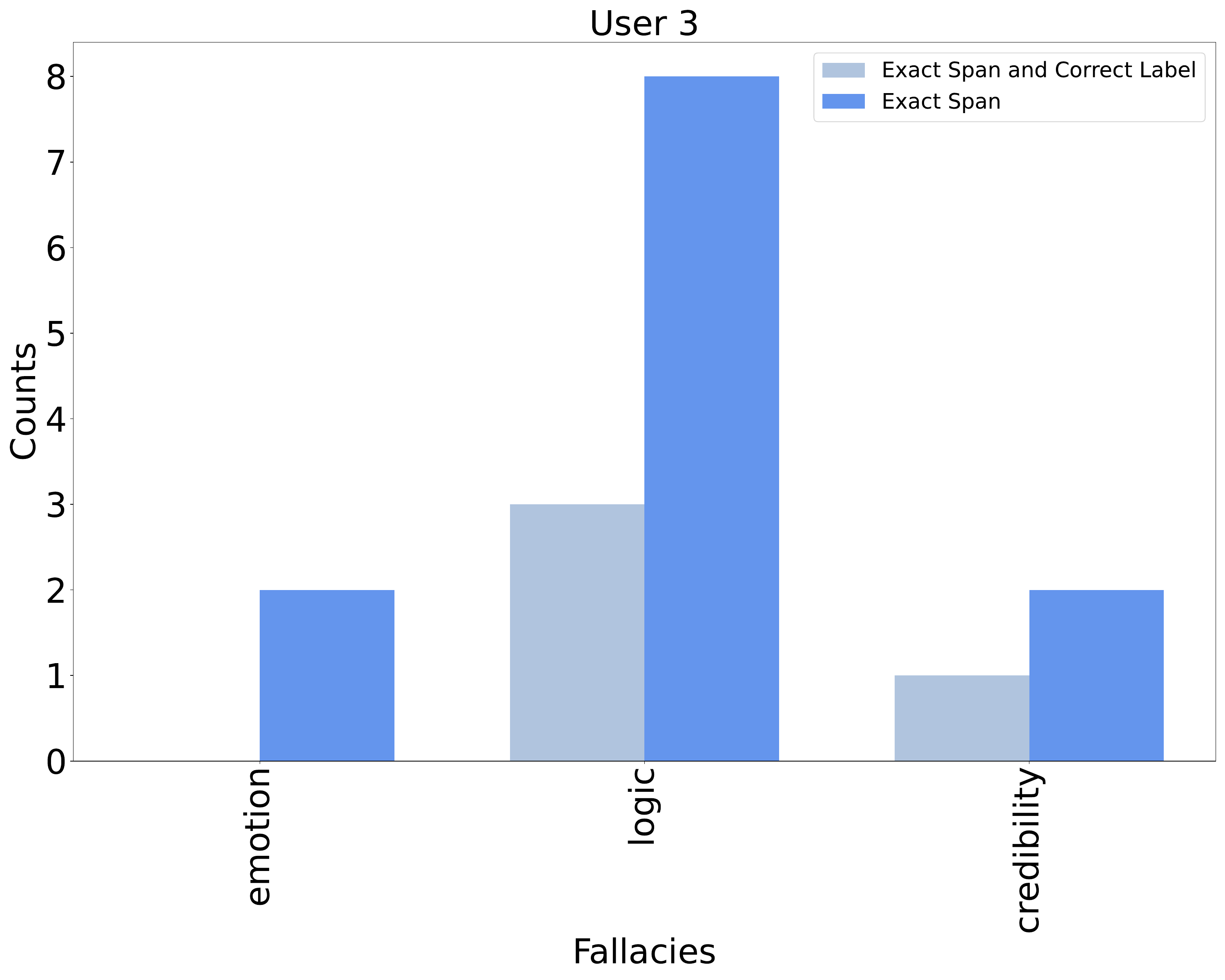}
        \caption{Accuracy of fallacy labeling for spans that exactly match the gold standard at Level 1 for the \textbf{User 3}}
        \label{fig:correct_fallacy_user_3_level_1}
    \end{subfigure}
    \hfill
    \begin{subfigure}[b]{0.45\textwidth}
        \includegraphics[width=\textwidth]{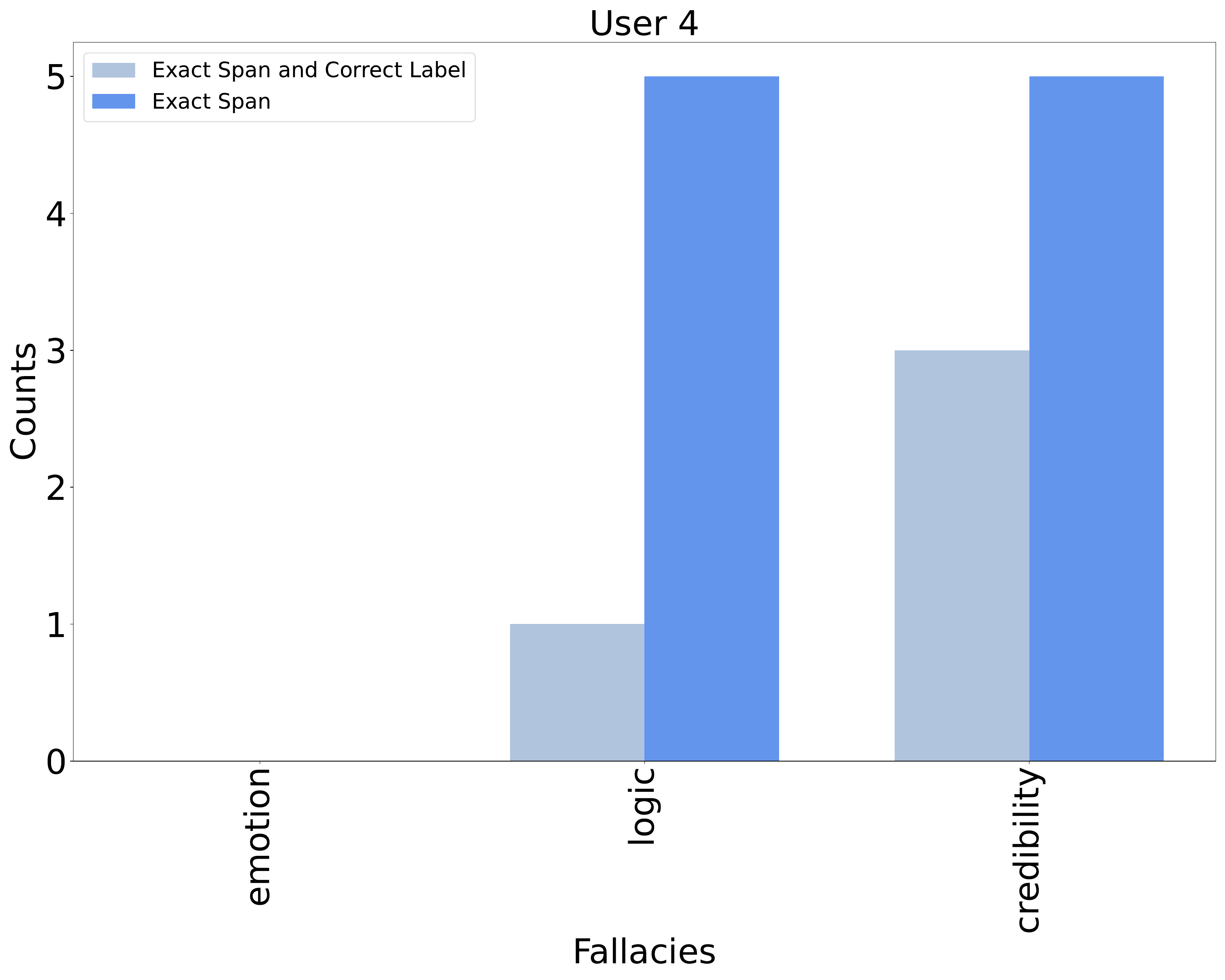}
        \caption{Accuracy of fallacy labeling for spans that exactly match the gold standard at Level 1 for the \textbf{User 4}}
        \label{fig:correct_fallacy_user_4_level_1}
    \end{subfigure}

    \caption{Accuracy of fallacy labeling for spans that \textbf{exactly match} the gold standard at Level 1 for the Users' annotations. \emph{Exact Span} corresponds to the number of spans correctly identified by the user, \emph{Exact Span and Correct Label} corresponds to the number of correctly labeled spans out of the correctly identified spans.}
    \label{fig:correct_fallacy_users_level_1}
\end{figure*}

We conduct an error analysis on two models, GPT-3.5 and Falcon, which exhibit the best and worst performance on Level 2. Our analysis also includes the annotations of the users study. \textbf{Our first goal in this analysis is to compare whether the best model has better controlled behavior than the worst model when generating outputs}. The Falcon model identifies 625 fallacious spans, with an average of 4.8 fallacies per span, while the GPT-3.5 model detects only 199 fallacious spans, with an average of 1.07 fallacies per span. However, we have 203 fallacious spans in the gold standard. 
The distribution of fallacies for the fallacious span at Level~2 for each model is presented in Table~\ref{tab:distribution_fallacies_level_2}. Based on our analysis, we have observed that the Falcon model tends to predict multiple fallacies that are irrelevant to a fallacious span.
In contrast, the GPT-3.5 model displays a more controlled behavior, which explains why Falcon has a low precision score. It is also worth noting that GPT-3.5 never predicted a span as \emph{tu quoque}.
We observe that both models produce nonsensical outputs, such as SQL code like ``\textit{select name color order from tag where the name},'' or incomplete classification of fallacies such as ``\textit{the sentence it's a mistake being considered as part of a fallacious argument.}''. Falcon has 115 spans labeled as unknown, while GPT-3.5 has only 5.  \textbf{Our second goal in this analysis is to analyze the exact matching performance of detecting fallacies and the type of fallacies that models and humans struggle with at Level~1}. Out of the 625 fallacious spans identified by Falcon, only 60 match the gold standard exactly, while out of 199 fallacious spans detected by GPT-3.5, only 55 match the gold standard exactly. Both models struggle mainly with fallacies categorized as fallacies of emotion, as shown in Figure~\ref{fig:correct_fallacy_models_level_1}.\\
For the annotators of the user study, we use a small sample of 20 examples with 24 spans. User 2 performs the best with 17 exactly matched spans, while User 4 performs worst with only 8 exactly matched spans. Based on the exact matched results, the analysis of Figure \ref{fig:correct_fallacy_users_level_1} reveals that all the annotators struggle mainly with the fallacies of 
 appeal to emotion. This difficulty can be partly attributed to these fallacies being less prevalent in our sample compared to the other types of fallacies. Interestingly, Users 1 and 3 correctly predict more fallacies of logic. Conversely, Users 2 and 4 correctly predict more fallacies of credibility than the others. It is worth noting that none of the users used all 23 fallacies of the taxonomy during the annotations, as shown in Table~\ref{tab:users_distribution_fallacies_level_2}.\\
In conclusion, models and humans tend to struggle more with fallacies of appeal to emotion, which could be expected since not every expression of emotion is necessarily a fallacy. The difficulty of the task lies in distinguishing valid arguments accompanied by emotions from fallacious arguments. 
This is supported by Figures~\ref{fig:correct_fallacy_models_level_1} and \ref{fig:correct_fallacy_users_level_1}. Despite the underrepresentation of the fallacies of the appeal to emotion in our user study sample, our findings indicate that humans often fail to exactly identify the specific fallacious spans classified under appeal to emotion fallacies. Moreover, even when humans correctly identify such fallacious spans, they are frequently misclassified. In contrast, models tend more to find these fallacious spans although they, too, frequently misclassify them. The only instances where the models can correctly predict the fallacious spans and their labels are when they involve an \emph{appeal to ridicule} or an \emph{appeal to a worse problem}. These cases can be observed in Figures~\ref{fig:correct_fallacy_models_level_2}.

\begin{table*}[ht]
\centering
\label{tab:fallacyComparison}
\begin{tabular}{@{}lccc@{}}
\toprule
Fallacy Type & Best Model & Worst Model & Gold Standard \\ \midrule
Appeal to Positive Emotion & 3 & 128 & 3 \\
Appeal to Anger & 6 & 119 & 6 \\
Appeal to Fear & 5 & 132 & 11 \\
Appeal to Pity & 1 & 198 & 1 \\
Appeal to Ridicule & 10 & 121 & 20 \\
Appeal to Worse Problems & 21 & 157 & 8 \\
Causal Oversimplification & 6 & 81 & 23 \\
Circular Reasoning & 8 & 132 & 11 \\
Equivocation & 1 & 106 & 7 \\
False Analogy & 6 & 127 & 12 \\
False Causality & 9 & 57 & 13 \\
False Dilemma & 6 & 169 & 18 \\
Hasty Generalization & 41 & 123 & 28 \\
Slippery Slope & 10 & 77 & 11 \\
Straw Man & 6 & 135 & 13 \\
Fallacy of Division & 2 & 102 & 2 \\
Ad Hominem & 32 & 135 & 16 \\
Ad Populum & 4 & 75 & 14 \\
Appeal to (False) Authority & 10 & 211 & 9 \\
Appeal to Nature & 7 & 143 & 11 \\
Appeal to Tradition & 4 & 156 & 6 \\
Guilt by Association & 4 & 91 & 8 \\
Tu Quoque & 0 & 111 & 3 \\
Unknown & 5 & 115 & - \\ \bottomrule
\end{tabular}
\caption{Fallacy distribution at Level 2 of the Gold standard, Best model and Worst model }
\label{tab:distribution_fallacies_level_2}
\end{table*}

\begin{table*}[ht]
\centering
\label{tab:fallacyComparison}
\begin{tabular}{@{}lccc@{}}
\toprule
Fallacy Type & Best Model & Worst Model & Gold Standard \\ \midrule
Emotion & 46 & 855 & 49 \\
Logic & 95 & 1109 & 138 \\
Credibility & 61 & 922 & 67 \\
Unknown & 5 & 115 & 0 \\ \bottomrule
\end{tabular}
\caption{Fallacy distribution at Level 1 of the Gold standard, Best model and Worst model }
\label{tab:distribution_fallacies_level_1}
\end{table*}

\begin{table*}[ht]
\centering

\begin{tabular}{@{}lccccc@{}}
\toprule
Fallacy Type & User 1 & User 2 & User 3 & User 4 & Sample Gold Standard \\ \midrule
Appeal to Positive Emotion & 2 & 0 & 0 & 2 & 0 \\
Appeal to Anger & 1 & 0 & 0 & 0 & 0 \\
Appeal to Fear & 1 & 1 & 0 & 2 & 0 \\
Appeal to Pity & 0 & 0 & 0 & 0 & 0 \\
Appeal to Ridicule & 8 & 1 & 1 & 4 & 5 \\
Appeal to Worse Problems & 3 & 0 & 0 & 0 & 1 \\
Causal Oversimplification & 2 & 2 & 1 & 4 & 2 \\
Circular Reasoning & 2 & 0 & 2 & 0 & 1 \\
Equivocation & 1 & 0 & 0 & 5 & 1 \\
False Analogy & 1 & 1 & 1 & 0 & 0 \\
False Causality & 3 & 4 & 2 & 1 & 2 \\
False Dilemma & 1 & 1 & 0 & 1 & 2 \\
Hasty Generalization & 4 & 2 & 3 & 5 & 3 \\
Slippery Slope & 1 & 1 & 0 & 7 & 1 \\
Straw Man & 2 & 5 & 0 & 0 & 3 \\
Fallacy of Division & 3 & 0 & 0 & 0 & 0 \\
Ad Hominem & 4 & 1 & 3 & 2 & 4 \\
Ad Populum & 3 & 1 & 0 & 5 & 1 \\
Appeal to (False) Authority & 0 & 2 & 1 & 3 & 1 \\
Appeal to Nature & 0 & 1 & 0 & 0 & 0 \\
Appeal to Tradition & 1 & 1 & 1 & 2 & 2 \\
Guilt by Association & 1 & 3 & 1 & 1 & 1 \\
Tu Quoque & 0 & 1 & 2 & 0 & 0 \\
Unknown & 0 & 0 & 0 & 0 & 0 \\ \bottomrule
\end{tabular}
\caption{Fallacies distribution at Level 2 of User 1, User 2, User 3, User 4, and the sample gold standard}\label{tab:users_distribution_fallacies_level_2}
\end{table*}

\begin{figure*}[ht]
    \centering
    \begin{subfigure}[b]{0.45\textwidth}
        \includegraphics[width=\textwidth]{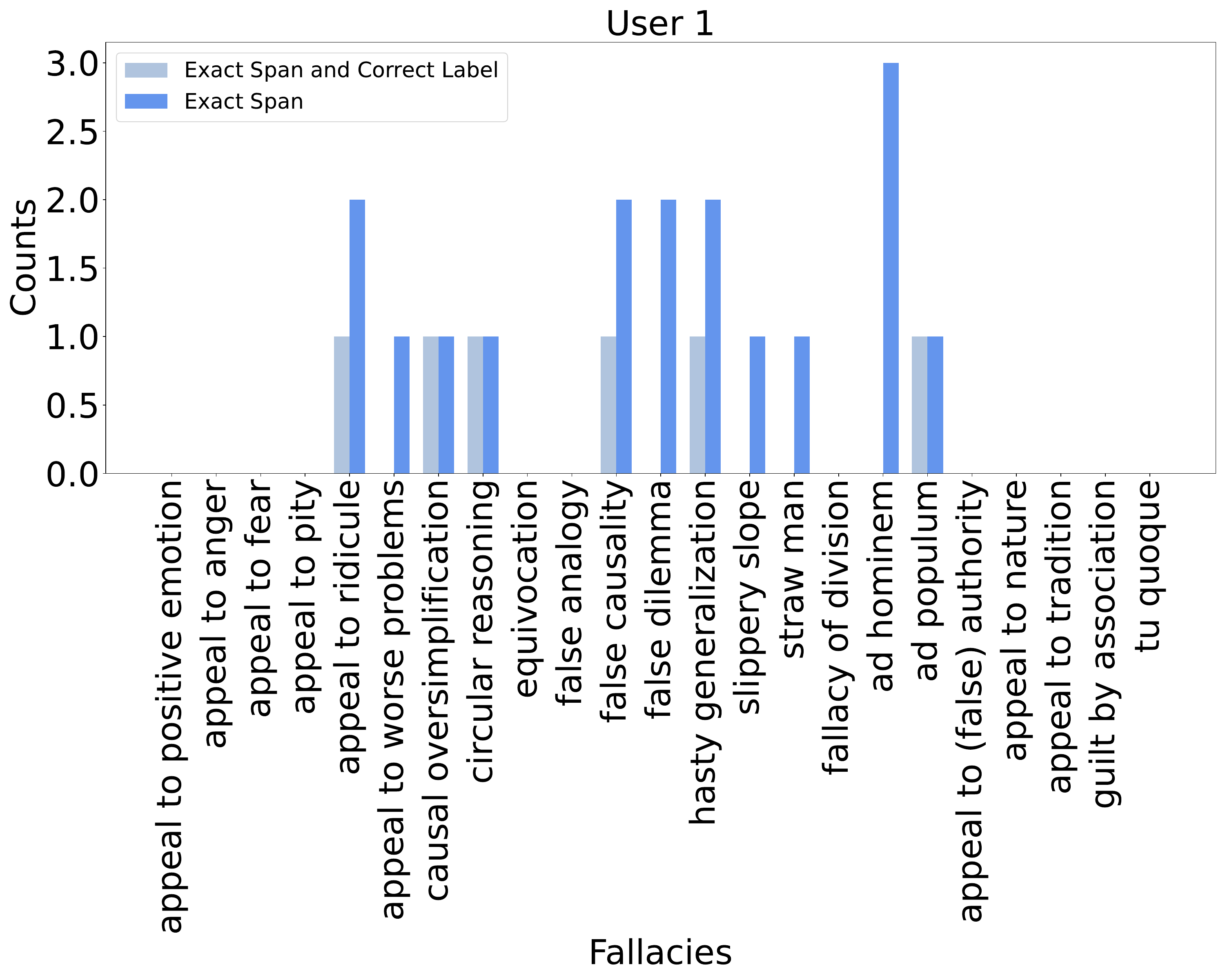}
        \caption{Accuracy of fallacy labeling for spans that exactly match the gold standard at Level 2 for the \textbf{User 1}}
        \label{fig:correct_fallacy_user_1_level_2}
    \end{subfigure}
    \hfill
    \begin{subfigure}[b]{0.45\textwidth}
        \includegraphics[width=\textwidth]{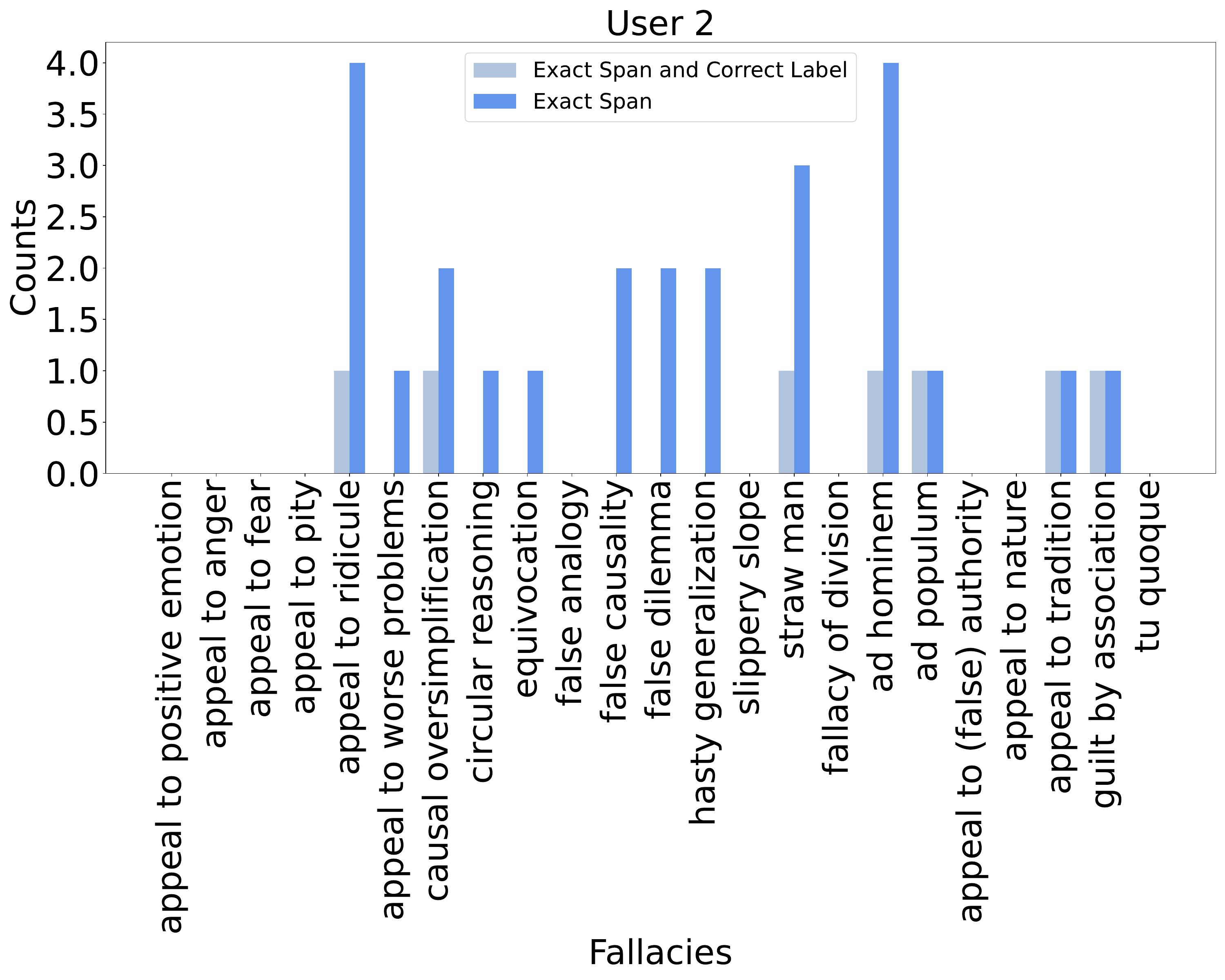}
        \caption{Accuracy of fallacy labeling for spans that exactly match the gold standard at Level 2 for the \textbf{User 2}}
        \label{fig:correct_fallacy_user_2_level_2}
    \end{subfigure}
    \begin{subfigure}[b]{0.45\textwidth}
        \includegraphics[width=\textwidth]{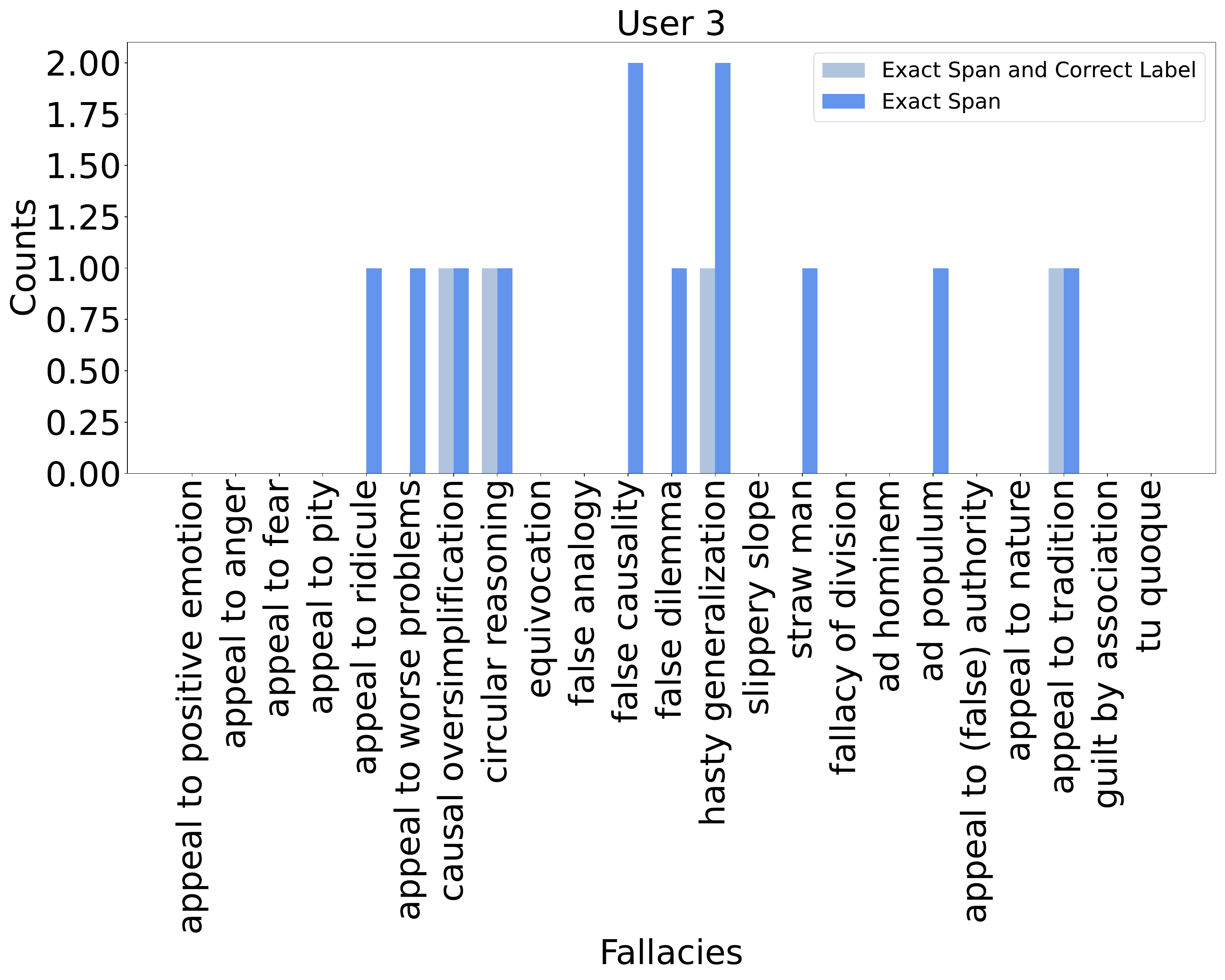}
        \caption{Accuracy of fallacy labeling for spans that exactly match the gold standard at Level 2 for the \textbf{User 3}}
        \label{fig:correct_fallacy_user_3_level_2}
    \end{subfigure}
    \hfill
    \begin{subfigure}[b]{0.45\textwidth}
        \includegraphics[width=\textwidth]{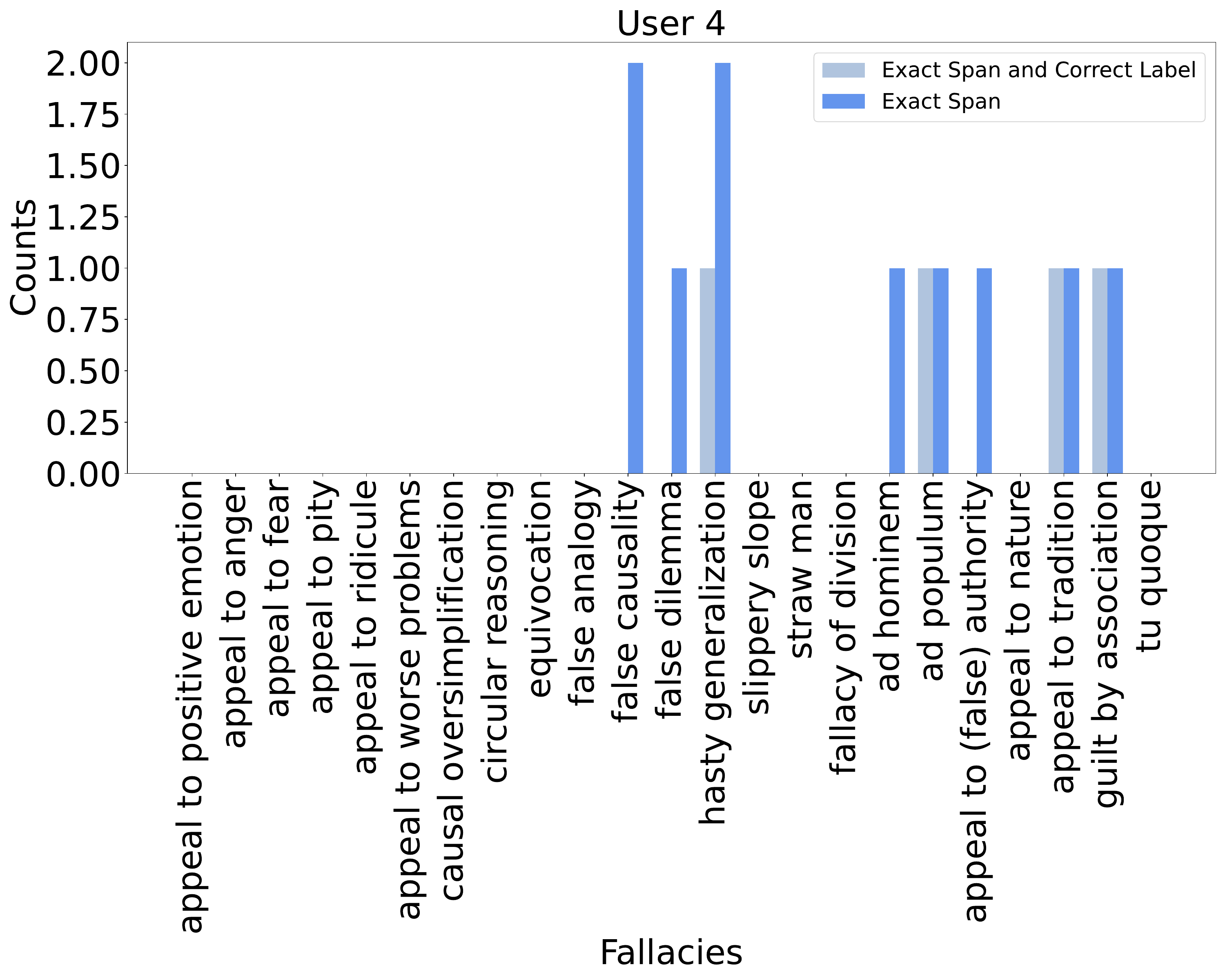}
        \caption{Accuracy of fallacy labeling for spans that exactly match the gold standard at Level 2 for the \textbf{User 4}}
        \label{fig:correct_fallacy_user_4_level_2}
    \end{subfigure}

    \caption{Accuracy of fallacy labeling for spans that \textbf{exactly match} the gold standard at Level 2 for the Users' annotations. \emph{Exact Span} corresponds to the number of spans correctly identified by the user, \emph{Exact Span and Correct Label} corresponds to the number of correctly labeled spans out of the correctly identified spans.}
    \label{fig:correct_fallacy_users_level_2}
\end{figure*}

\onecolumn
\section{Edge Cases of the Metrics} 

\label{app:metrics_edge_cases}
In this section, we show how our metrics handle edge cases of our disjunctive annotation scheme.

\begin{table}[H]
    \centering
    \caption{The model predicts at least one correct label}
    \begin{tabular}{llccc}
    \hline
    & Spans & Labels & & \\
    \hline
    \multirow{3}{*}{Gold} & $\text{\color{example1}{Lorem ipsum dolor sit amet.}}$ & $l1, \bot$ & & \\
    & $\text{\color{example2}{Ut enim ad minim veniam.}}$ & $l2$ & & \\
    & $\text{\color{example4}{Sed do eiusmod tempor incididunt.}}$ & $l3$ & & \\
    \hline
    \hline
    Case & Prediction & Label & Recall & Precision \\
    \hline
    \multirow{1}{*}{0.1} & $\text{\color{example2}Ut enim ad minim veniam.}$ & $l2$ & \multirow{1}{*}{0.5} & \multirow{1}{*}{1} \\
    \hline
    \multirow{2}{*}{0.2} & $\text{\color{example1}{Lorem ipsum dolor sit amet.}}$ & $l1$ & \multirow{2}{*}{0.5} & \multirow{2}{*}{1} \\
    & $\text{\color{example2}{Ut enim ad minim veniam.}}$ & $l2$ & & \\
    \hline
    \multirow{2}{*}{0.3} & $\text{\color{example2}{Ut enim ad minim veniam.}}$ & $l2$ & \multirow{2}{*}{0.5} & \multirow{2}{*}{0.5} \\
    & $\text{\color{example4}{Sed do eiusmod tempor incididunt.}}$ & $l4$ & & \\
    \hline
    \multirow{3}{*}{0.4} & $\text{\color{example1}{Lorem ipsum dolor sit amet.}}$ & $l1$ & \multirow{3}{*}{0.5} & \multirow{3}{*}{0.666} \\
     &  $\text{\color{example2}{Ut enim ad minim veniam.}}$ & $l2$  &  &  \\
     &  $\text{\color{example4}{Sed do eiusmod tempor incididunt.}}$ & $l4$  &  &  \\
    \hline

    \end{tabular}
    \label{tab:Prediction 0}
\end{table}

    \begin{table}[H]
        \centering
        \caption{The gold standard has only one span, which contains a ``no fallacy'' as an alternative}
        \begin{tabular}{llccc}
        \hline
        & Spans & Labels & Recall & Precision \\
        \hline
        \multirow{1}{*}{Gold} & $\text{\color{example1}{Lorem ipsum dolor sit amet.}}$ & $l1, \bot$ & & \\
        \hline
        \hline
        Case & Prediction & Label &  &  \\
        \hline
        \multirow{1}{*}{1.1} & $\text{\color{example1}Lorem ipsum dolor sit amet.}$ & $l1$ & \multirow{1}{*}{1} & \multirow{1}{*}{1} \\
        \hline
        \multirow{1}{*}{1.2} & $\text{\color{example1}{Lorem ipsum dolor sit amet.}}$ & $l3$ & \multirow{1}{*}{1} & \multirow{1}{*}{0} \\
        \hline
        \multirow{1}{*}{1.3} & - & - & \multirow{1}{*}{1} & \multirow{1}{*}{1} \\
        \hline
        \multirow{1}{*}{1.4} & $\text{\color{example2}{Ut enim ad minim veniam.}}$ & $l1$ & \multirow{1}{*}{1} & \multirow{1}{*}{0} \\
        \hline
        \multirow{1}{*}{1.5} & $\text{\color{example1}{Lorem ipsum dolor sit amet.} \color{example2}{Ut enim ad minim veniam.}}$ & $l3$ & \multirow{1}{*}{1} & \multirow{1}{*}{0} \\
        \hline

        \end{tabular}
        \label{tab:Prediction 1}
    \end{table}

    \begin{table}[H]
        \centering
        \caption{The gold standard does not contain a ``no fallacy''}
        \begin{tabular}{llccc}
        \hline
        & Spans & Labels & Recall & Precision \\
        \hline
        \multirow{1}{*}{Gold} & $\text{\color{example1}{Lorem ipsum dolor sit amet.}}$ & $l1$ & & \\
        \hline
        \hline
        Case & Prediction & Label &  &  \\
        \hline
        \multirow{1}{*}{2.1} & $\text{\color{example1}Lorem ipsum dolor sit amet.}$ & $l1$ & \multirow{1}{*}{1} & \multirow{1}{*}{1} \\
        \hline
        \multirow{1}{*}{2.2} & $\text{\color{example1}{Lorem ipsum dolor sit amet.}}$ & $l3$ & \multirow{1}{*}{0} & \multirow{1}{*}{0} \\
        \hline
        \multirow{1}{*}{2.3} & $\text{\color{example2}{Ut enim ad minim veniam.}}$ & $l1$ & \multirow{1}{*}{0} & \multirow{1}{*}{0} \\
        \hline
        \multirow{1}{*}{2.4} & - & - & \multirow{1}{*}{0} & \multirow{1}{*}{1} \\
        \hline

        \end{tabular}
        \label{tab:Prediction 2}
    \end{table}

    \begin{table}[H]
        \centering
        \caption{The gold standard contains no fallacious span}
        \begin{tabular}{llccc}
        \hline
        & Spans & Labels & Recall & Precision \\
        \hline
        \multirow{1}{*}{Gold} & - & - & & \\
        \hline
        \hline
        Case & Prediction & Label &  &  \\
        \hline
        \multirow{1}{*}{3.1} & $\text{\color{example1}Lorem ipsum dolor sit amet.}$ & $l1$ & \multirow{1}{*}{1} & \multirow{1}{*}{0} \\
        \hline
        \multirow{1}{*}{3.2} & - & - & \multirow{1}{*}{1} & \multirow{1}{*}{1} \\
        \hline

        \end{tabular}
        \label{tab:Prediction 3}
    \end{table}

     \begin{table}[H]
        \centering
        \caption{The gold standard contains a ``no fallacy'' and a required fallacy}
        \begin{tabular}{llccc}
        \hline
        & Spans & Labels & Recall & Precision \\
        \hline
        \multirow{2}{*}{Gold} & $\text{\color{example1}{Lorem ipsum dolor sit amet.}}$ & $l1, \bot$ & & \\
        & $\text{\color{example2}{Ut enim ad minim veniam.}}$ & $l2$ & & \\
        \hline
        \hline
        Case & Prediction & Label &  &  \\
        \hline
        \multirow{1}{*}{4.1} & $\text{\color{example1}Lorem ipsum dolor sit amet.}$ & $l1$ & \multirow{1}{*}{0} & \multirow{1}{*}{1} \\
        \hline
        \multirow{1}{*}{4.2} & $\text{\color{example1}{Lorem ipsum dolor sit amet.}}$ & $l3$ & \multirow{1}{*}{0} & \multirow{1}{*}{0} \\
        \hline
        \multirow{1}{*}{4.3} & $\text{\color{example2}{Ut enim ad minim veniam.}}$ & $l2$ & \multirow{1}{*}{1} & \multirow{1}{*}{1} \\
        \hline
        \multirow{1}{*}{4.4} & $\text{\color{example2}{Ut enim ad minim veniam.}}$ & $l3$ & \multirow{1}{*}{0} & \multirow{1}{*}{0} \\
        \hline

        \end{tabular}
        \label{tab:Prediction 4}
    \end{table}

    \begin{table}[H]
        \centering
        \caption{The gold standard spans across 2 sentences}
        \begin{tabular}{llccc}
        \hline
        & Spans & Labels & Recall & Precision \\
        \hline
        \multirow{1}{*}{Gold} & $\text{\color{example1}{Lorem ipsum dolor sit amet.} \color{example2}{Ut enim ad minim veniam.}}$ & $l1, \bot$ & & \\
        \hline
        \hline
        Case & Prediction & Label &  &  \\
        \hline
        \multirow{1}{*}{5.1} & $\text{\color{example1}Lorem ipsum dolor sit amet. \color{example2}Ut enim ad minim veniam.}$ & $l1$ & \multirow{1}{*}{1} & \multirow{1}{*}{1} \\
        \hline
        \multirow{1}{*}{5.2} & - & - & \multirow{1}{*}{1} & \multirow{1}{*}{1} \\
        \hline
        \multirow{1}{*}{5.3} & $\text{\color{example1}{Lorem ipsum dolor sit amet.}}$ & $l1$ & \multirow{1}{*}{1} & \multirow{1}{*}{1} \\
        \hline
        \multirow{1}{*}{5.4} & $\text{\color{example1}{Lorem ipsum dolor sit amet.} \color{example2}{Ut enim ad minim veniam.}}$ & $l3$ & \multirow{1}{*}{1} & \multirow{1}{*}{0} \\
        \hline
        \multirow{1}{*}{5.5} & $\text{\color{example2}{Ut enim ad minim veniam.}}$ & $l3$ & \multirow{1}{*}{1} & \multirow{1}{*}{0} \\
        \hline

        \end{tabular}
        
        \label{tab:Prediction 5}
    \end{table}

     \begin{table}[H]
        \centering
        \caption{The gold standard spans across 2 sentences and there is overlap}
        \begin{tabular}{llccc}
        \hline
        & Spans & Labels & Recall & Precision \\
        \hline
        \multirow{2}{*}{Gold} & $\text{\color{example1}{Lorem ipsum dolor sit amet.} \color{example2}{Ut enim ad minim veniam.}}$ & $l1, \bot$ & & \\
         & $\text{\color{example2}{Ut enim ad minim veniam.}}$ & $l2$ & & \\
        \hline
        \hline
        Case & Prediction & Label &  &  \\
        \hline
        \multirow{2}{*}{6.1} & $\text{\color{example1}{Lorem ipsum dolor sit amet.} \color{example2}{Ut enim ad minim veniam.}}$ & $l1$ & \multirow{2}{*}{1} & \multirow{2}{*}{1} \\
         & $\text{\color{example2}{Ut enim ad minim veniam.}}$ & $l2$ &  &  \\
        \hline
        \multirow{1}{*}{6.2} & $\text{\color{example2}{Ut enim ad minim veniam.}}$ & $l2$ & \multirow{1}{*}{1} & \multirow{1}{*}{1} \\
        \hline
        \multirow{1}{*}{6.3} & $\text{\color{example1}{Lorem ipsum dolor sit amet.} \color{example2}{Ut enim ad minim veniam.}}$ & $l1$ & \multirow{1}{*}{0} & \multirow{1}{*}{1} \\
        \hline
        \multirow{1}{*}{6.4} & - & - & \multirow{1}{*}{0} & \multirow{1}{*}{1} \\
        \hline
        \multirow{2}{*}{6.5} & $\text{\color{example1}{Lorem ipsum dolor sit amet.}}$ & $l1$ & \multirow{2}{*}{1} & \multirow{2}{*}{1} \\
         & $\text{\color{example2}{Ut enim ad minim veniam.}}$ & $l2$ &  &  \\
        \hline
        \multirow{1}{*}{6.6} & $\text{\color{example1}{Lorem ipsum dolor sit amet.}}$ & $l1$ & \multirow{1}{*}{0} & \multirow{1}{*}{1} \\
        \hline
        \multirow{1}{*}{6.7} & $\text{\color{example2}{Ut enim ad minim veniam.}}$ & $l3$ & \multirow{1}{*}{0} & \multirow{1}{*}{0} \\
        \hline

        \end{tabular}
        \label{tab:Prediction 6}
    \end{table}

    \begin{table}[H]
        \centering
        \caption{Two gold standard spans overlap}
        \begin{tabular}{llccc}
        \hline
        & Spans & Labels & Recall & Precision \\
        \hline
        \multirow{2}{*}{Gold} & $\text{\color{example1}{Lorem ipsum dolor sit amet.} \color{example2}{Ut enim ad minim veniam.}}$ & $l1$ & & \\
         & $\text{\color{example2}{Ut enim ad minim veniam.}}$ & $l2$ & & \\
        \hline
        \hline
        Case & Prediction & Label &  &  \\
        \hline
        \multirow{2}{*}{7.1} & $\text{\color{example1}{Lorem ipsum dolor sit amet.} \color{example2}{Ut enim ad minim veniam.}}$ & $l1$ & \multirow{2}{*}{1} & \multirow{2}{*}{1} \\
         & $\text{\color{example2}{Ut enim ad minim veniam.}}$ & $l2$ &  &  \\
        \hline
        \multirow{1}{*}{7.2} & $\text{\color{example2}{Ut enim ad minim veniam.}}$ & $l2$ & \multirow{1}{*}{0.5} & \multirow{1}{*}{1} \\
        \hline
        \multirow{1}{*}{7.3} & $\text{\color{example1}{Lorem ipsum dolor sit amet.} \color{example2}{Ut enim ad minim veniam.}}$ & $l1$ & \multirow{1}{*}{0.5} & \multirow{1}{*}{1} \\
        \hline
        \multirow{1}{*}{7.4} & - & - & \multirow{1}{*}{0} & \multirow{1}{*}{1} \\
        \hline
        \multirow{2}{*}{7.5} & $\text{\color{example1}{Lorem ipsum dolor sit amet.}}$ & $l1$ & \multirow{2}{*}{0.75} & \multirow{2}{*}{1} \\
         & $\text{\color{example2}{Ut enim ad minim veniam.}}$ & $l2$ &  &  \\
        \hline
        \multirow{1}{*}{7.6} & $\text{\color{example1}{Lorem ipsum dolor sit amet.}}$ & $l1$ & \multirow{1}{*}{0.25} & \multirow{1}{*}{1} \\
        \hline
        \multirow{1}{*}{7.7} & $\text{\color{example2}{Ut enim ad minim veniam.}}$ & $l3$ & \multirow{1}{*}{0} & \multirow{1}{*}{0} \\
        \hline

        \end{tabular}
        \label{tab:Prediction 7}
    \end{table}

    \begin{table}[H]
        \centering
        \caption{Two labels have the same Level 0 or Level 1 fallacy category}
        \begin{tabular}{llccc}
        \hline
        & Spans & Labels & Recall & Precision \\
        \hline
        \multirow{2}{*}{Gold} & $\text{\color{example1}{Lorem ipsum dolor sit amet.} \color{example2}{Ut enim ad minim veniam.}}$ & fallacy (l1) & & \\
         & $\text{\color{example2}{Ut enim ad minim veniam.}}$ & fallacy (l2) & & \\
        \hline
        \hline
        Case & Prediction & Label &  &  \\
        \hline
        \multirow{2}{*}{8.1} & $\text{\color{example1}{Lorem ipsum dolor sit amet.} \color{example2}{Ut enim ad minim veniam.}}$ & fallacy & \multirow{2}{*}{1} & \multirow{2}{*}{1} \\
         & $\text{\color{example2}{Ut enim ad minim veniam.}}$ & fallacy &  &  \\
        \hline
        \multirow{1}{*}{8.2} & $\text{\color{example2}{Ut enim ad minim veniam.}}$ & fallacy & \multirow{1}{*}{0.75} & \multirow{1}{*}{1} \\
        \hline
        \multirow{1}{*}{8.3} & $\text{\color{example1}{Lorem ipsum dolor sit amet.} \color{example2}{Ut enim ad minim veniam.}}$ & fallacy & \multirow{1}{*}{1} & \multirow{1}{*}{1} \\
        \hline
        \multirow{2}{*}{8.4} & $\text{\color{example2}{Ut enim ad minim veniam.}}$ & fallacy & \multirow{2}{*}{0.75} & \multirow{2}{*}{0.5} \\
         & $\text{\color{example2}{Ut enim ad minim veniam.}}$ & fallacy (duplicate) &  &  \\
        \hline
        \multirow{2}{*}{8.5} & $\text{\color{example1}{Lorem ipsum dolor sit amet.}}$ & fallacy & \multirow{2}{*}{0.75} & \multirow{2}{*}{1} \\
         & $\text{\color{example2}{Ut enim ad minim veniam.}}$ & fallacy &  &  \\
        \hline
        \multirow{1}{*}{8.6} & $\text{\color{example1}{Lorem ipsum dolor sit amet.}}$ & fallacy (l1) & \multirow{1}{*}{0.25} & \multirow{1}{*}{1} \\
        \hline
        \multirow{1}{*}{8.7} & $\text{\color{example1}{Lorem ipsum dolor sit amet.}}$ & fallacy (l2) & \multirow{1}{*}{0.25} & \multirow{1}{*}{1} \\
        \hline

        \end{tabular}
        \label{tab:Prediction 8}
    \end{table}

    \begin{table}[H]
        \centering
        \caption{Two labels have the same Level 0 or Level 1 fallacy category with an alternative ``no fallacy''}
        \begin{tabular}{llccc}
        \hline
        & Spans & Labels & Recall & Precision \\
        \hline
        \multirow{2}{*}{Gold} & $\text{\color{example2}{Ut enim ad minim veniam.}}$ & fallacy (l1), $\bot$ & & \\
         & $\text{\color{example2}{Ut enim ad minim veniam.}}$ & fallacy (l2) & & \\
        \hline
        \hline
        Case & Prediction & Label &  &  \\
        \hline
        \multirow{2}{*}{9.1} & $\text{\color{example2}{Ut enim ad minim veniam.}}$ & fallacy & \multirow{2}{*}{1} & \multirow{2}{*}{1} \\
         & $\text{\color{example2}{Ut enim ad minim veniam.}}$ & fallacy (duplicate) &  &  \\
        \hline
        \multirow{1}{*}{9.2} & $\text{\color{example2}{Ut enim ad minim veniam.}}$ & fallacy (l1) & \multirow{1}{*}{1} & \multirow{1}{*}{1} \\
        \hline
        \multirow{1}{*}{9.3} & $\text{\color{example2}{Ut enim ad minim veniam.}}$ & fallacy (l2) & \multirow{1}{*}{1} & \multirow{1}{*}{1} \\
        \hline
        \multirow{1}{*}{9.4} & $\text{\color{example1}{Lorem ipsum dolor sit amet.} \color{example2}{Ut enim ad minim veniam.}}$ & fallacy & \multirow{1}{*}{1} & \multirow{1}{*}{0.5} \\
        \hline
        \multirow{1}{*}{9.5} & - & - & \multirow{1}{*}{0} & \multirow{1}{*}{1} \\
        \hline
        \multirow{1}{*}{9.6} & $\text{\color{example1}{Lorem ipsum dolor sit amet.}}$ & fallacy & \multirow{1}{*}{0} & \multirow{1}{*}{0} \\
        \hline

        \end{tabular}
        \label{tab:Prediction 9}
    \end{table}

    \begin{table}[H]
        \centering
        \caption{The same obligatory fallacious span has different labels}
        \begin{tabular}{llccc}
        \hline
        & Spans & Labels & Recall & Precision \\
        \hline
        \multirow{2}{*}{Gold} & $\text{\color{example1}{Lorem ipsum dolor sit amet.}}$ & $l1$ & & \\
         & $\text{\color{example1}{Lorem ipsum dolor sit amet.}}$ & $l2$ & & \\
        \hline
        \hline
        Case & Prediction & Label &  &  \\
        \hline
        \multirow{1}{*}{10.1} & $\text{\color{example1}{Lorem ipsum dolor sit amet.}}$ & $l1$ & \multirow{1}{*}{0.5} & \multirow{1}{*}{1} \\
        \hline
        \multirow{1}{*}{10.2} & $\text{\color{example1}{Lorem ipsum dolor sit amet.}}$ & $l3$ & \multirow{1}{*}{0} & \multirow{1}{*}{0} \\
        \hline
        \multirow{1}{*}{10.3} & $\text{\color{example2}{Ut enim ad minim veniam.}}$ & $l1$ & \multirow{1}{*}{0} & \multirow{1}{*}{0} \\
        \hline
        \multirow{1}{*}{10.4} & - & - & \multirow{1}{*}{0} & \multirow{1}{*}{1} \\
        \hline
        \multirow{2}{*}{10.5} & $\text{\color{example1}{Lorem ipsum dolor sit amet.}}$ & $l1$ & \multirow{2}{*}{1} & \multirow{2}{*}{1} \\
         & $\text{\color{example1}{Lorem ipsum dolor sit amet.}}$ & $l2$ &  &  \\
        \hline

        \end{tabular}
        \label{tab:Prediction 10}
    \end{table}

    \begin{table}[H]
        \centering
        \caption{The same fallacious span has two labels and a ``no fallacy'' alternative}
        \begin{tabular}{llccc}
        \hline
        & Spans & Labels & Recall & Precision \\
        \hline
        \multirow{2}{*}{Gold} & $\text{\color{example1}{Lorem ipsum dolor sit amet.}}$ & $l1, \bot$ & & \\
         & $\text{\color{example1}{Lorem ipsum dolor sit amet.}}$ & $l2$ & & \\
        \hline
        \hline
        Case & Prediction & Label &  &  \\
        \hline
        \multirow{1}{*}{11.1} & $\text{\color{example1}{Lorem ipsum dolor sit amet.}}$ & $l1$ & \multirow{1}{*}{1} & \multirow{1}{*}{1} \\
        \hline
        \multirow{1}{*}{11.2} & $\text{\color{example1}{Lorem ipsum dolor sit amet.}}$ & $l3$ & \multirow{1}{*}{0} & \multirow{1}{*}{0} \\
        \hline
        \multirow{1}{*}{11.3} & $\text{\color{example2}{Ut enim ad minim veniam.}}$ & $l1$ & \multirow{1}{*}{0} & \multirow{1}{*}{0} \\
        \hline
        \multirow{1}{*}{11.4} & - & - & \multirow{1}{*}{0} & \multirow{1}{*}{1} \\
        \hline
        \multirow{2}{*}{11.5} & $\text{\color{example1}{Lorem ipsum dolor sit amet.}}$ & $l1$ & \multirow{2}{*}{1} & \multirow{2}{*}{1} \\
         & $\text{\color{example1}{Lorem ipsum dolor sit amet.}}$ & $l2$ &  &  \\
        \hline

        \end{tabular}
        \label{tab:Prediction 11}
    \end{table}

\end{document}